\documentclass[runningheads]{llncs}

\usepackage[accsupp]{axessibility}  

\usepackage[table]{xcolor}
\usepackage{color}

\usepackage{float}       
\usepackage{subcaption}  
\usepackage{multirow}
\usepackage{graphicx}
\usepackage{comment}
\usepackage{amsmath}
\usepackage{amssymb}
\usepackage{booktabs}
\usepackage{multirow}

\usepackage{algorithmic}
\usepackage{algorithm}
\usepackage{wrapfig}
\usepackage{listings}
\usepackage{lipsum}

\usepackage{tabularx}
\usepackage{makecell}
\usepackage{verbatim}
\usepackage{diagbox}

\usepackage{amssymb}
\usepackage{pifont}

\usepackage{mathtools}

\usepackage{etoolbox}
\makeatletter
\AfterEndEnvironment{algorithm}{\let\@algcomment\relax}
\AtEndEnvironment{algorithm}{\kern2pt\hrule\relax\vskip3pt\@algcomment}
\let\@algcomment\relax
\newcommand\algcomment[1]{\def\@algcomment{\footnotesize#1}}
\renewcommand\fs@ruled{\def\@fs@cfont{\bfseries}\let\@fs@capt\floatc@ruled
  \def\@fs@pre{\hrule height.8pt depth0pt \kern2pt}%
  \def\@fs@post{}%
  \def\@fs@mid{\kern2pt\hrule\kern2pt}%
  \let\@fs@iftopcapt\iftrue}
\makeatother

\usepackage[pagebackref,breaklinks,colorlinks]{hyperref}

\usepackage[capitalize]{cleveref}
\crefname{section}{Sec.}{Secs.}
\Crefname{section}{Section}{Sections}
\Crefname{table}{Table}{Tables}
\crefname{table}{Tab.}{Tabs.}

\def\ie{\textit{i.e.,~}}

\def\etc{\textit{etc~}}

\def\sota{state-of-the-art~}

\usepackage{amsmath,amsfonts,bm}

\def\eqref#1{equation~\ref{#1}}

\def\1{\bm{1}}

\newcommand{\test}{\mathcal{D_{\mathrm{test}}}}

\def\mA{{\bm{A}}}

\def\mG{{\bm{G}}}

\def\mK{{\bm{K}}}

\def\mQ{{\bm{Q}}}

\def\mV{{\bm{V}}}

\def\mX{{\bm{X}}}
\def\mY{{\bm{Y}}}
\def\mZ{{\bm{Z}}}

\DeclareMathAlphabet{\mathsfit}{\encodingdefault}{\sfdefault}{m}{sl}
\SetMathAlphabet{\mathsfit}{bold}{\encodingdefault}{\sfdefault}{bx}{n}

\begin{document}
\pagestyle{headings}
\mainmatter
\def\ECCVSubNumber{7935}  

\title{Eliminating Gradient Conflict in Reference-based Line-Art Colorization} 

\titlerunning{Eliminating Gradient Conflict in Reference-based Line-art Colorization}
%
\author{Zekun Li\inst{1} \and
Zhengyang Geng\inst{2} \and
Zhao Kang\inst{1}\thanks{Corresponding author}\and
Wenyu Chen\inst{1} \and
Yibo Yang\inst{3}}
\authorrunning{Li et al.}
%
\institute{University of Electronic Science and Technology of China, Chengdu, China \\  \email{kunkun0w0@std.uestc.edu.cn;\{zkang,cwy\}@uestc.edu.cn}\\
\and
Peking University, School of AI, Beijing, China \\
\email{ZhengyangGeng@gmail.com}\\
\and
JD Explore Academy \\ \email{ibo@pku.edu.cn}}
\maketitle

\begin{abstract}
Reference-based line-art colorization is a challenging task in computer vision.
The color, texture, and shading are rendered based on an abstract sketch, which heavily relies on the precise long-range dependency modeling between the sketch and reference.
Popular techniques to bridge the cross-modal information and model the long-range dependency employ the attention mechanism.
However, in the context of reference-based line-art colorization, several techniques would intensify the existing training difficulty of attention, for instance, self-supervised training protocol and GAN-based losses.
To understand the instability in training, we detect the gradient flow of attention and observe gradient conflict among attention branches.
This phenomenon motivates us to alleviate the gradient issue by preserving the dominant gradient branch while removing the conflict ones. 
We propose a novel attention mechanism using this training strategy, Stop-Gradient Attention (SGA), outperforming the attention baseline by a large margin with better training stability.
Compared with \sota modules in line-art colorization, our approach demonstrates significant improvements in Fr$\text{\'{e}}$chet Inception Distance (FID, up to 27.21\%) and structural similarity index measure (SSIM, up to 25.67\%) on several benchmarks. The code of SGA is available at \href{https://github.com/kunkun0w0/SGA}{https://github.com/kunkun0w0/SGA}.

\end{abstract}

\section{Introduction}
Reference-based line-art colorization has achieved impressive performance in generating a realistic color image from a line-art image \cite{graph-anime-colorization,cocosnet2}.
This technique is in high demand in comics, animation, and other content creation applications \cite{aiqiyi-cmft,animation-transformer}.
Different from painting with other conditions such as color strokes \cite{s2p5,DCSGAN}, palette \cite{palette}, or text \cite{tag2pix}, using a style reference image as condition input not only provides richer semantic information for the model but also eliminates the requirements of precise color information and the geometric hints provided by users for every step.
Nevertheless, due to the huge information discrepancy between the sketch and reference, it is challenging to correctly transfer colors from reference to the same semantic region in the sketch.

\begin{figure}
    \centering
    \includegraphics[width=0.6\linewidth]{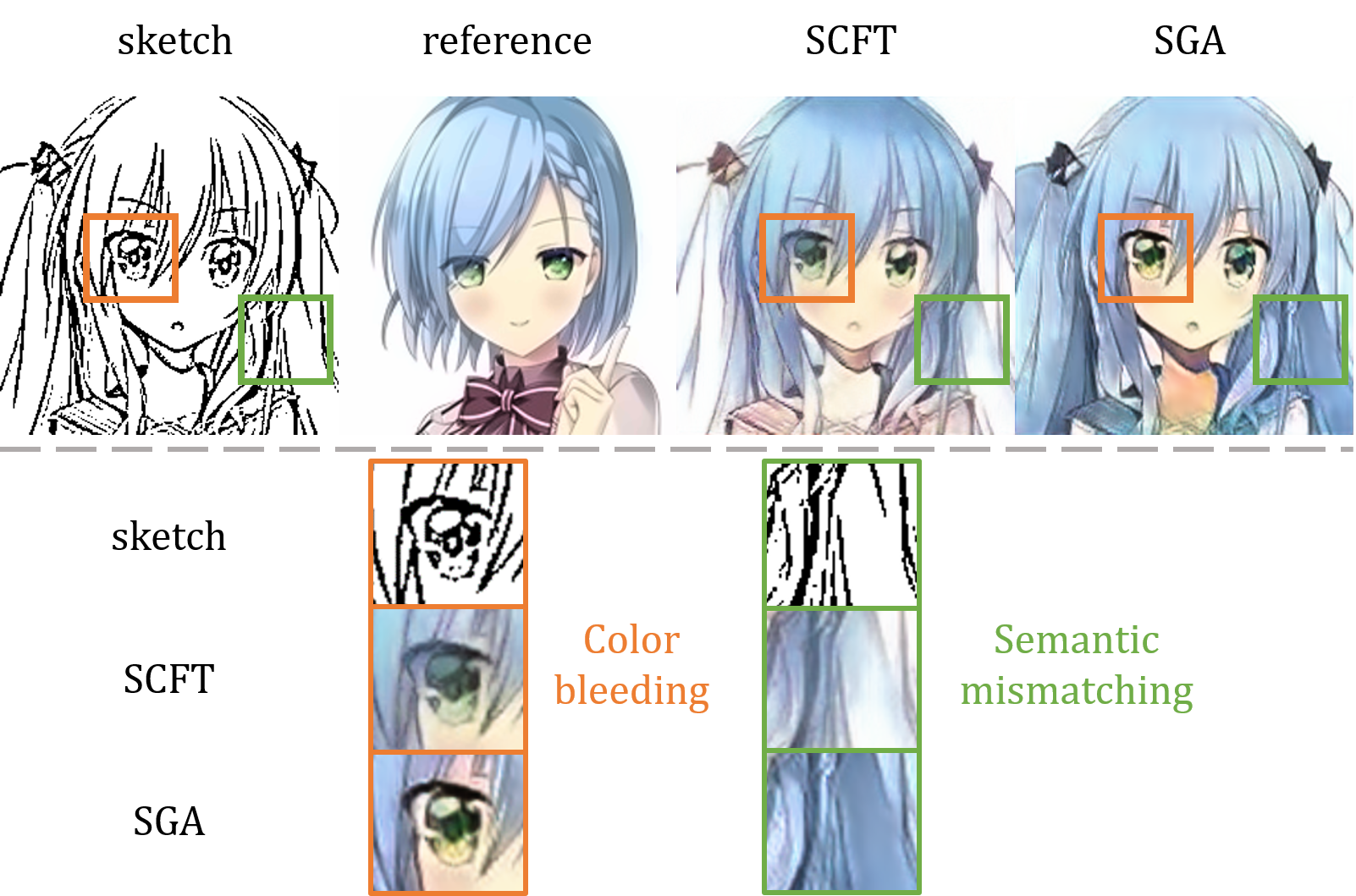}
    \caption{
    The comparison between the images produced by SCFT \cite{scft} and SGA (Ours).
    SCFT subjects to \textcolor[RGB]{237,125,49}{color bleeding (orange box)} and \textcolor[RGB]{112,173,71}{semantic mismatching (green box)}.
    }
    \label{fig:compare-page1}
\end{figure}

Several methods attempt to tackle the reference-based colorization by fusing the style latent code of reference into the sketch \cite{s2p,icon,spade}.
Inspired by the success of the attention mechanism~\cite{Attention_is_All_you_Need,non_local_net}, researchers adopt attention modules to establish the semantic correspondence and inject colors by mapping the reference to the sketch \cite{scft,aiqiyi-cmft,cocosnet}.
However, as shown in \Cref{fig:compare-page1}, the images generated by these methods often contain color bleeding or semantic mismatching, indicating considerable room for improving attention methods in line-art colorization.

There are many possible reasons for the deficiency of line-art colorization using attention: model pipeline, module architecture, or training.
Motivated by recent works \cite{ham,mocov3} concerning the training issues of attention models, we are particularly interested in the training stability of attention modules in line-art colorization.
It is even more challenging to train attention models in line-art colorization because state-of-the-art models \cite{scft} deploy multiple losses using a GAN-style training pipeline, which can double the training instability.
Therefore, we carefully analyze the training dynamics of attention in terms of its gradient flow in the context of line-art colorization.
We observe the gradient conflict phenomenon, namely, a gradient branch contains a negative cosine similarity with the summed gradient.

To eliminate the gradient conflict, we detach the conflict one while preserving the dominant gradient, which ensures that the inexact gradient has a positive cosine similarity with the exact gradient and meet theory requirements \cite{Geng2021Training,pcgrad}.
This training strategy visibly boosts the training stability and performance compared with the baseline attention colorization models.
Combined with architecture design, this paper introduces \textbf{S}top-\textbf{G}radient \textbf{A}ttention, \textbf{SGA}, whose training strategy eliminates the gradient conflict and helps the model learn better colorization correspondence.
SGA properly transfers the style of reference images to the sketches, establishing accurate semantic correspondence between sketch-reference pairs.
Our experiment results on several image domains show clear improvements over previous methods, \ie up to 27.21\% and 25.67\% regarding FID and SSIM, respectively.

Our contributions are summarized as follows:
\begin{itemize}
\setlength{\itemsep}{2pt}
    \item We reveal the gradient conflict in attention mechanism for line-art colorization, \ie a gradient branch contains a negative cosine similarity with the summed gradient.
    \item We propose a novel attention mechanism with gradient and design two attention blocks based on SGA, \ie cross-SGA and self-SGA.
    \item Both quantitative and qualitative results verify that our method outperforms state-of-the-art modules on several image datasets.
\end{itemize}

\section{Related Work}
\subsubsection{Reference-based Line-Art Colorization.} 
The reference-based line-art colorization is a user-friendly approach to assist designers in painting the sketch with their desired color~\cite{s2p,scft,aiqiyi-cmft,animation-transformer}.
Early studies attempt to get the style latent code of reference and directly mix it with sketch feature maps to generate the color image \cite{s2p,icon}.
To make better use of reference images, some studies propose spatial-adaptive normalization methods~\cite{spade,SEAN}.

Different from the aforementioned methods that adopt latent vectors for style control, \cite{scft,aiqiyi-cmft,cocosnet} learn dense semantic correspondences between sketch-reference pairs.
These approaches utilize the dot-product attention \cite{Attention_is_All_you_Need,non_local_net} to model the semantic mapping between sketch-reference pairs and inject color into sketch correctly.
Although traditional non-local attention is excellent in feature alignment and integration between different modalities, the model cannot learn robust representation due to the gradient conflict in attention's optimization.
Thus, our work proposes the stop-gradient operation for attention to eliminate the gradient conflict problem in line-art colorization.

\subsubsection{Attention Mechanism.} 
The attention mechanism~\cite{visual_attention,Attention_is_All_you_Need} is proposed to capture long-range dependencies and align signals from different sources. It is widely applied in vision~\cite{non_local_net,SAGAN}, language~\cite{Attention_is_All_you_Need,TransformerXL}, and graph~\cite{GAT} areas. 
Due to the quadratic memory complexity of standard dot-product attention, many researchers from the vision \cite{A^2-Nets,GloRe,emanet,latentGNN,ham} and language~\cite{TransformersAreRNN,Linformer,RoutingT} communities endeavor to reduce the memory consumption to linear complexity. 
Recently, vision transformer~\cite{ViT} starts a new era for modeling visual data through the attention mechanism.
The booming researches using transformer substantially change the trend in image~\cite{Deit,Swin,PVT,CvT}, point cloud~\cite{PCT,PT}, gauge~\cite{he2021gauge}, and video~\cite{VTN,ViViT} processing.

Unlike existing works concerning the architectures of attention mechanism, we focus on the training of attention modules regarding its gradient flow. Although some strategies have been developed to improve the training efficiency \cite{Deit,SAMViT} for vision transformer, they mainly modify the objective function to impose additional supervision. From another perspective, our work investigates the gradient issue in the attention mechanism.

\subsubsection{Stop-Gradient Operation.}
Backpropagation is the foundation for training deep neural networks. 
Recently some researchers have paid attention to the gradient flow in the deep models.
Hamburger~\cite{ham} proposes the one-step gradient to tackle the gradient conditioning and gradient norm issues in the implicit global context module, which helps obtain stable learning and performance.
SimSiam~\cite{simsiam} adopts the one-side stop-gradient operation to implicitly introduce an extra set of variables to implement Expectation-Maximization (EM) like algorithm in contrastive learning.
VQ-VAE~\cite{VQVAE} also encourages discrete codebook learning by the stop-gradient supervision.
All of these works indicate the indispensability of the gradient manipulation, which demonstrates that the neural network performance is related to both the advanced architecture and the appropriate training strategy.

Inspired by prior arts, our work investigates the gradient conflict issue for training non-local attention.
The stop-gradient operation clips the conflict gradient branches while preserving correction direction for model updates.

\section{Proposed Method}
\subsection{Overall Workflow}\label{sec:overall}

\begin{figure*}[ht]
\centering
\includegraphics[width=0.9\linewidth]{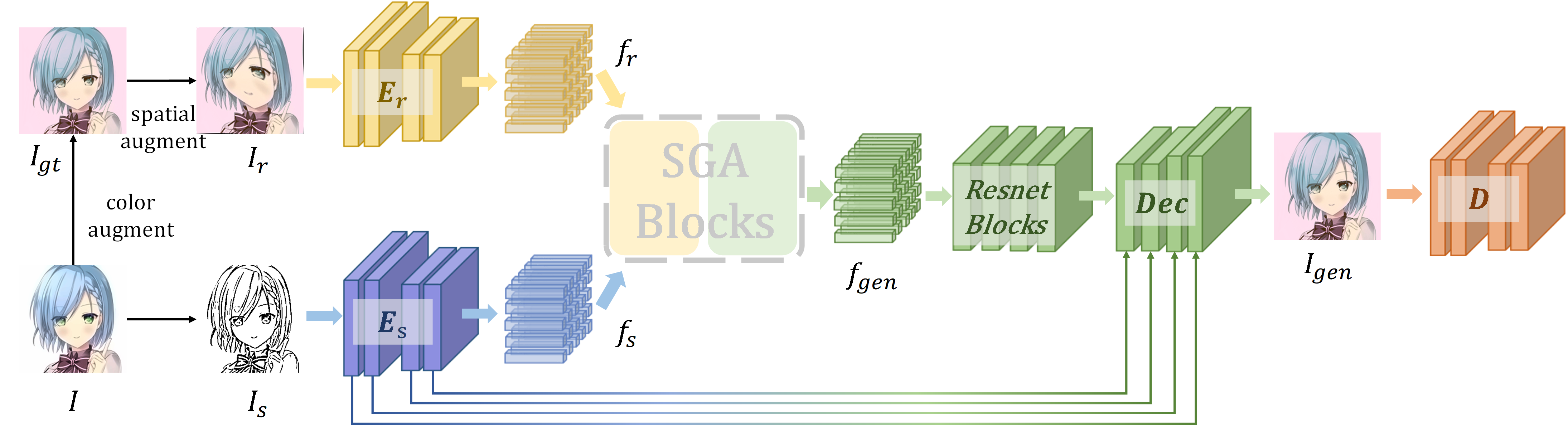}
\caption{
The overview of our reference-based line-art colorization framework with a discriminator $\boldsymbol{D}$: 
Given the sketch $I_s$, the target image $I_{gt}$ and the reference $I_r$ obtained through original image $I$, we input $I_s$ and $I_r$ into encoder $E_s$ and $E_r$ to extract feature maps $f_s$ and $f_r$.
The SGA blocks, which contain \textcolor[RGB]{191,144,0}{cross-SGA} and \textcolor[RGB]{84,130,53}{self-SGA}, integrate $f_{s}$ and $f_{r}$ into the mixed feature map $f_{gen}$.
Then $f_{gen}$ is passed through several residual blocks and a U-net decoder $Dec$ with skip connection to generate the image $I_{gen}$.
The $I_{gen}$ is supposed to be similar to $I_{gt}$. 
}
\label{fig:generator}
\end{figure*}

As illustrated in \cref{fig:generator}, we adopt a self-supervised training process similar to \cite{scft}.
Given a color image $I$, we first use XDoG \cite{2012xdog} to convert it into a line-art image $I_s$.
Then, the expected coloring result $I_{gt}$ is obtained by adding a random color jittering on $I$ .
Additionally, we generate a style reference image $I_{r}$ through applying the thin plate splines transformation on $I_{gt}$.

In the training process, utilizing $I_{r}$ as the reference to color the sketch $I_{s}$, our model first uses encoder $E_s$ and $E_r$ to extract sketch feature $f_s \in \mathbb{R}^{c \times h \times w}$ and reference feature $f_r \in \mathbb{R}^{c \times h \times w}$.
In order to leverage multi-level representation simultaneously for feature alignment and integration, we concatenate the feature maps of all convolution layers outputs after using 2D adaptive average pooling function to down-sample them into the same spatial size. 

To integrate the content in sketch and the style in reference, we employ our SGA blocks.
There are two types of SGA blocks in our module: \textcolor[RGB]{191,144,0}{cross-SGA} integrates the features from different domains and \textcolor[RGB]{84,130,53}{self-SGA} models the global context of input features.
Then several residual blocks and a U-net decoder $Dec$ with skip connections to sketch encoder $E_{s}$ are adopted to generate the image $I_{gen}$ by the mixed feature map $f_{gen}$.
In the end, we add an adversarial loss~\cite{gan} by using a discriminator $\boldsymbol{D}$ to distinguish the output $I_{gen}$ and the ground truth~$I_{gt}$.

\subsection{Loss Function}
\noindent\textbf{Image Reconstruction Loss.}
\quad According to the Section~\ref{sec:overall}, both generated images $I_{gen}$ and ground truth images $I_{gt}$ should keep style consistency with reference $I_r$ and outline preservation with sketch $I_s$.
Thus, we adopt $L_1$ regularization to measure the difference between $I_{gen}$ and $I_{gt}$, which ensures that the model colors correctly and distinctly:
\begin{equation}
    \mathcal{L}_{\text {rec }}=\mathbb{E}_{I_{s},I_{r},I_{gt}}\left[\left\|\mG\left(I_{s}, I_{r}\right)-I_{g t}\right\|_{1}\right]
    \label{eq:l1-image}
\end{equation}
where $\mG \left(I_{s}, I_{r}\right)$ means coloring the sketch $I_s$ with the reference $I_r$.

~\\
\noindent\textbf{Adversarial Loss.}
\quad In order to generate a realistic image with the same outline as the prior sketch $I_{s}$, we leverage a conditional discriminator $\boldsymbol{D}$ to distinguish the generated images from real ones~\cite{isola2017image}.
The least square adversarial loss~\cite{lsgan} for optimizing our GAN-based model is formulated as:
\begin{equation}
    \mathcal{L}_{a d v} =\mathbb{E}_{I_{g t}, I_{s}}\left[\left\|\boldsymbol{D}\left(I_{g t}, I_{s}\right)\right\|_{2}^2\right] +\mathbb{E}_{I_{s}, I_{r}}\left[\left\|\left(\mathbf{1}-\boldsymbol{D}\left(\boldsymbol{G}\left(I_{s}, I_{r}\right), I_{s}\right)\right)\right\|_{2}^2\right]
\label{eq:adv-loss}
\end{equation}

~\\
\noindent\textbf{Style and Perceptual Loss.}
\quad As shown in previous works~\cite{scft,Perceptual_loss}, perceptual loss and style loss encourage a network to produce a perceptually plausible output.
Leveraging the ImageNet pretrained network, we reduce the gaps in multi-layer activation outputs between the target image $I_{gt}$ and generated image $I_{gen}$ by minimizing the following losses:
\begin{align}
    \mathcal{L}_{perc}&=\mathbb{E}_{I_{gt},I_{gen}}\left[\sum_{l}\left\|\phi_{l}(I_{gt})-\phi_{l}\left(I_{gen}\right)\right\|_{1}\right] \\
    \mathcal{L}_{\text {style }}&=\mathbb{E}_{I_{gt},I_{gen}}\left[\left\|\mathcal{G}\left(\phi_{l}(I_{gt})\right)-\mathcal{G}\left(\phi_{l}\left(I_{gen}\right)\right)\right\|_{1}\right]
\end{align}
where $\phi_l$ represents the activation map of the $l_{th}$ layer extracted at the relu from VGG19 network, and $\mathcal{G}$ is the gram matrix.

~\\
\noindent\textbf{Overall Loss}
\quad In summary, the overall loss function for the generator $\boldsymbol{G}$ and discriminator $\boldsymbol{D}$ is defined as:
\begin{equation}
    \min _{\boldsymbol{G}} \max _{\boldsymbol{D}} \mathcal{L}_{\text{total}} =\mathcal{L}_{a d v}+\lambda_{1} \mathcal{L}_{\text {rec}} + \lambda_{2} \mathcal{L}_{\text {perc}}+\lambda_{3} \mathcal{L}_{\text {style}}
    \label{eq:sum-loss}
\end{equation}

\subsection{Gradient Issue in Attention}
\label{sec:discussion}

In this section, we use SCFT~\cite{scft}, a classic attention-based method in colorization, as an example to study the gradient issue in attention.
$\mQ \in \mathbb{R}^{n \times d}$ is the feature projection transformed by $\mathbf{W}_{\mathclap{\rm q}}$\, from the input $\mX \in \mathbb{R}^{n \times d}$.
The feature projections $\mK,\mV \in \mathbb{R}^{n \times d}$ from input $\mY \in\mathbb{R}^{n \times d}$ are transformed by $\mathbf{W}_{\mathclap{\rm k}}$\, and $\mathbf{W}_{\mathclap{\rm v}}$ .
Given the attention map $\mA \in \mathbb{R}^{n \times n}$, the classic dot-product attention mechanism can be formulated as follows:
\begin{equation}
    \mZ = \text{softmax}(\frac{\mQ \mK^\top}{\sqrt{d}})\mV + \mX = \mA \mV + \mX
    \label{eq:self-attention}
\end{equation}

\begin{figure}[ht]
    \centering
    \begin{subfigure}{0.3\textwidth}
    \includegraphics[width=0.9\linewidth]{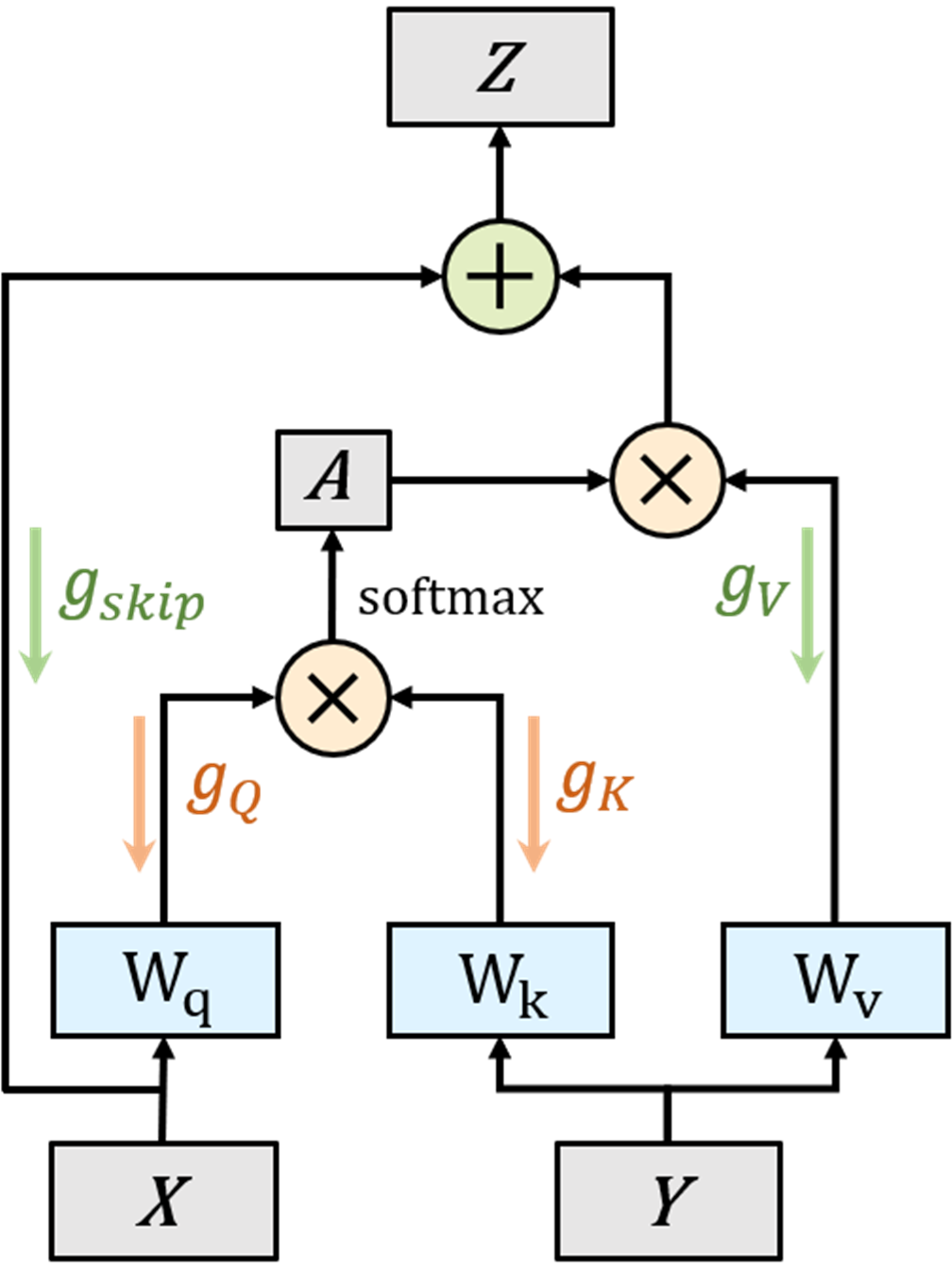}
    \caption{SCFT}
    \label{fig:scft-module}
    \end{subfigure}
    \hspace{0.02\textwidth}
    \begin{subfigure}{0.3\textwidth}
    \includegraphics[width=0.9\linewidth]{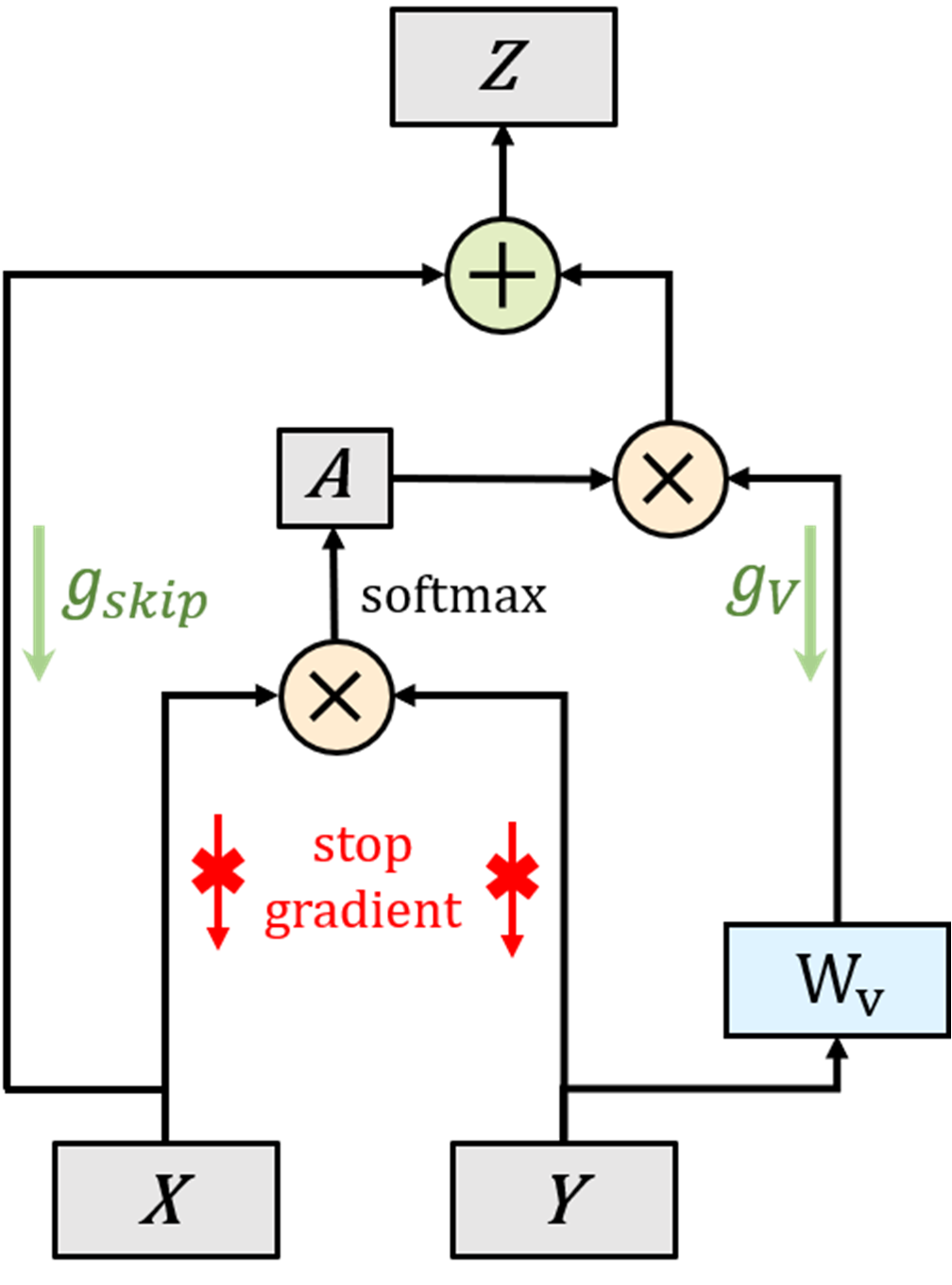}
    \caption{SCFT w/ stop-grad}
    \label{fig:scft-wo-g-module}
    \end{subfigure}
    \caption{
    Stop-gradient in attention module.
    The $\boldsymbol{g}_{skip}$, $\boldsymbol{g}_{Q}$, $\boldsymbol{g}_{K}$ and $\boldsymbol{g}_{V}$ separately represent the gradient along their branches.
    The stop-gradient operation (\textbf{stop-grad}) truncates the backpropagation of conflict gradients existing in attention map calculation.
    }
    \label{fig:scft&sga-page3}
\end{figure}

Previous works~\cite{mocov3,ham,SAMViT,Deit} present the training difficulty of vision attention: instability, worse generalization, \etc. For line-art colorization, it is even more challenging to train the attention models, as the training involves GAN-style loss and reconstruction loss, which are understood to lead to mode collapse~\cite{gan} or trivial solutions. Given a training schedule, the loss of colorization network can shake during training and finally deteriorate.

\begin{figure*}
    \centering
    \begin{subfigure}{0.45\textwidth}
    \includegraphics[width=0.9\linewidth]{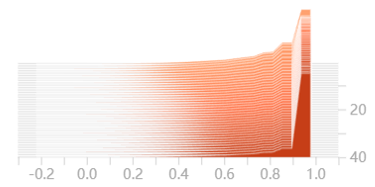}
    \caption{$\cos{(\boldsymbol{g}_{skip},\boldsymbol{g}_{skip} + \boldsymbol{g}_{\boldsymbol{Q}})}$ in branch $\mX$}
    \label{fig:v-skt}
    \end{subfigure}
    \begin{subfigure}{0.45\textwidth}
    \includegraphics[width=0.9\linewidth]{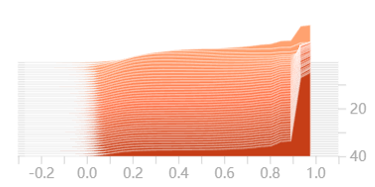}
    \caption{$\cos{(\boldsymbol{g}_{\boldsymbol{V}},\boldsymbol{g}_{\boldsymbol{K}} + \boldsymbol{g}_{\boldsymbol{V}})}$ in branch $\mY$}
    \label{fig:v-ref}
    \end{subfigure}
    \\
    \begin{subfigure}{0.45\textwidth}
    \includegraphics[width=0.9\linewidth]{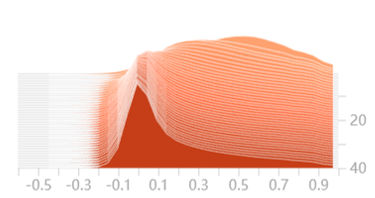}
    \caption{$\cos{(\boldsymbol{g}_{\boldsymbol{Q}},\boldsymbol{g}_{skip} + \boldsymbol{g}_{\boldsymbol{Q}})}$ in branch $\mX$}
    \label{fig:a-skt}
    \end{subfigure}
    \begin{subfigure}{0.45\textwidth}
    \includegraphics[width=0.9\linewidth]{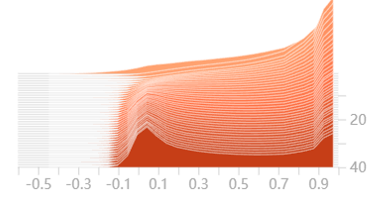}
    \caption{$\cos{(\boldsymbol{g}_{\boldsymbol{K}},\boldsymbol{g}_{\boldsymbol{K}} + \boldsymbol{g}_{\boldsymbol{V}})}$ in branch $\mY$}
    \label{fig:a-ref}
    \end{subfigure}
    \caption{
    The histograms of the gradient cosine value distribution in 40 epochs.
    A large cosine value means that the network mainly uses this branch of gradient to optimize the loss function.
    }
    \label{fig:gradient}
\end{figure*}

To better understand reasons behind the training difficulty of attention in colorization, we analyze the gradient issue through the classic SCFT model \cite{scft}. We visualize the gradient flow back through the attention module in terms of each gradient branch and the summed gradient. 

\cref{fig:gradient} offers the cosine value between different gradient branches and the total gradient.
We separately calculate 
$\cos{(\boldsymbol{g}_{skip},\boldsymbol{g}_{skip} + \boldsymbol{g}_{\boldsymbol{Q}})}$ 
and 
$\left.\mathrm{cos}(\boldsymbol{g}_{\boldsymbol{Q}},\boldsymbol{g}_{skip} \notag\right. \\ \left.+ \boldsymbol{g}_{\boldsymbol{Q}}) \right.$ 
for each pixel in branch $\mX$ (means gradient in sketch feature maps $f_s$), 
$\cos{(\boldsymbol{g}_{\boldsymbol{V}},\boldsymbol{g}_{\boldsymbol{K}} + \boldsymbol{g}_{\boldsymbol{V}})}$ and 
$\cos{(\boldsymbol{g}_{\boldsymbol{K}},\boldsymbol{g}_{\boldsymbol{K}} + \boldsymbol{g}_{\boldsymbol{V}})}$ in branch $\mY$ (means gradient in reference feature maps $f_r$) 
to explore the gradient flow of the network during learning.

Note that first order optimization methods usually require the surrogate gradient $\Tilde{\boldsymbol{g}}$ for update to be ascent, \ie $\cos{(\Tilde{\boldsymbol{g}}, \boldsymbol{g})} > 0$,
where $\boldsymbol{g}$ is the exact gradient. Then the update direction based on the surrogate gradient can be descent direction. The visualization in \cref{fig:gradient} implies that the gradient $\boldsymbol{g}_{skip}$ from the skip connection for the branch $\mX$ and the gradient $\boldsymbol{g}_{V}$ from $\mV$ for the branch $\mY$ has already become an ascent direction for optimization, denoting that $\boldsymbol{g}_{Q}$ and $\boldsymbol{g}_{K}$ from the attention map construct the ``conflict gradient'' $\not {\boldsymbol{g}}$ in respect of the total gradient $\boldsymbol{g}$, \ie
$\cos{(\not {\boldsymbol{g}}, \boldsymbol{g})} < 0$.

\cref{fig:v-skt,fig:v-ref} show that $\boldsymbol{g}_{skip}$ and $\boldsymbol{g}_{V}$ are usually highly correlated with the total gradient, where over \textbf{78.09\%} and \textbf{52.39\%} of the cosine values are greater than \textbf{0.935} in the 40th epoch, respectively.
Moreover, these percentages increase during training, indicating the significance of the representative gradient.
On the other hand, nearly \textbf{30.57\%} of $\boldsymbol{g}_{q}$ in \cref{fig:a-skt} and \textbf{10.77\%} of $\boldsymbol{g}_{K}$ in \cref{fig:a-ref} have negative cosine values in the 40th epoch.
These proportions are \textbf{22.81\%} and \textbf{5.32\%} in the 20th epoch, respectively, gradually increasing during training.

The visualization regarding the gradient flows demonstrates that the two gradient branches compete with each other for a dominant position during training process, while $\boldsymbol{g}_{skip}$ and $\boldsymbol{g}_{V}$ construct an ascent direction and $\boldsymbol{g}_{Q}$ and $\boldsymbol{g}_{K}$ remain as the conflict gradient in respect of the total gradient in each branch.
According to existing works in multi-task learning~\cite{pcgrad}, large gradient conflict ratios may result in significant performance drop.
It motivates us to detach the conflict gradient while preserving the dominant gradient as inexact gradient to approximate the original gradient, illustrated in \cref{fig:scft&sga-page3}. 

Verified by \cref{fig:v-skt,fig:v-ref}, the gradient after the stop-gradient operation forms an ascent direction of the loss landscape, \ie $\cos{(\Tilde{\boldsymbol{g}}, \boldsymbol{g})} > 0$, and thus be valid for optimization~\cite{Geng2021Training}.

\begin{wraptable}{l}{0.65\textwidth}
\centering
\caption{
Test the Fr$\text{\'{e}}$chet Inception Distance (FID) and SSIM with different settings of SCFT on anime dataset.
$\uparrow$ means the higher the better, while $\downarrow$ indicates the lower the better. 
}
\begin{tabular}{cccc}
\Xhline{1pt} \specialrule{0em}{0pt}{1pt}
\multicolumn{2}{c}{SCFT\, Setting} &  \multirow{2}{*}{\quad FID$\downarrow$ \quad}  & \multirow{2}{*}{\quad SSIM$\uparrow$ \quad}    \\ 
\cline{1-2} \specialrule{0em}{0pt}{1.5pt}
\quad \textbf{stop-grad} \quad & \quad $\mathbf{W}_{\mathclap{\rm q}}$ \&$\mathbf{W}_{\mathclap{\rm k}}$ \quad &\\
\Xhline{1pt} \specialrule{0em}{0pt}{1pt}
\ding{55} & \ding{51}  & 44.65   & 0.788 \\ 
\hline \specialrule{0em}{0pt}{1pt}
\ding{55} & \ding{55}    & 48.04 & 0.799       \\ 
\hline \specialrule{0em}{0pt}{1pt}
\ding{51} & \ding{51}   & 38.20 &0.835 \\
\hline \specialrule{0em}{0pt}{1pt}
\ding{51} & \ding{55}       & \cellcolor[HTML]{DAE8FC}\textbf{36.78} & \cellcolor[HTML]{DAE8FC}\textbf{0.841} \\ 
\Xhline{1pt}
\end{tabular}
\label{tab:scft-ab}
~\\
~\\
~\\
~\\
\end{wraptable}

Table \ref{tab:scft-ab} shows that the gradient clipping through the stop-gradient operation can effectively improve the model performance. 
We can also remove $\mathbf{W}_{\mathclap{\rm k}}$\,  and $\mathbf{W}_{\mathclap{\rm q}}$ since there is no gradient propagating in them and they will not be updated in the training process.
The lower FID and higher SSIM mean that model can generate more realistic images with higher outline preservation during colorization after the stop-gradient clipping. 

\begin{figure}[ht]
\centering
\includegraphics[width=0.7\textwidth]{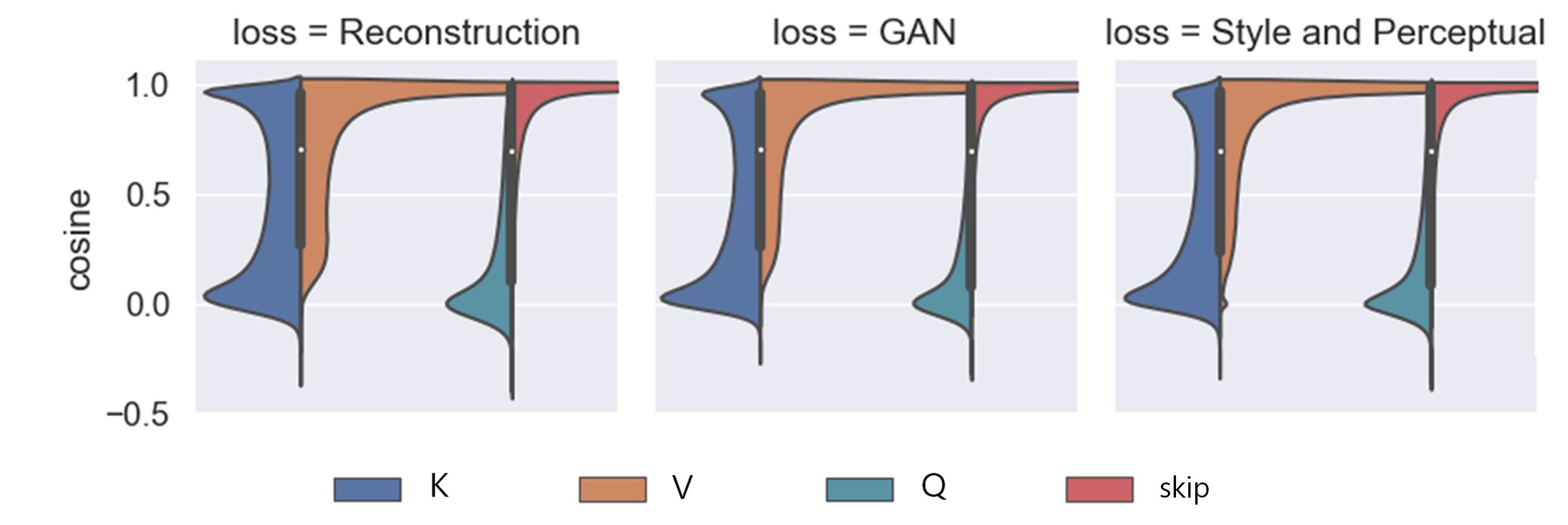}
\caption{Visualizations of the gradient cosine distribution when using a single loss on the pretrained SCFT model.}
\label{fig:ab-loss-grad-distributions}
\end{figure}

In order to investigate the reliability of gradient conflicts, we test the gradient cosine distributions when using a certain loss to confirm the trigger to gradient issue is the dot-product attention.
We use the SCFT model to compute the gradients cosine distribution of each loss to investigate whether loss functions or architectures cause the conflict.
\cref{fig:ab-loss-grad-distributions} shows that all loss terms cause similar conflicts, implying that the attention architecture leads to gradient conflicts.

\subsection{Stop-Gradient Attention}
\begin{figure}[ht]
\centering
\begin{subfigure}{0.445\textwidth}
\centering
\includegraphics[width=0.71\linewidth]{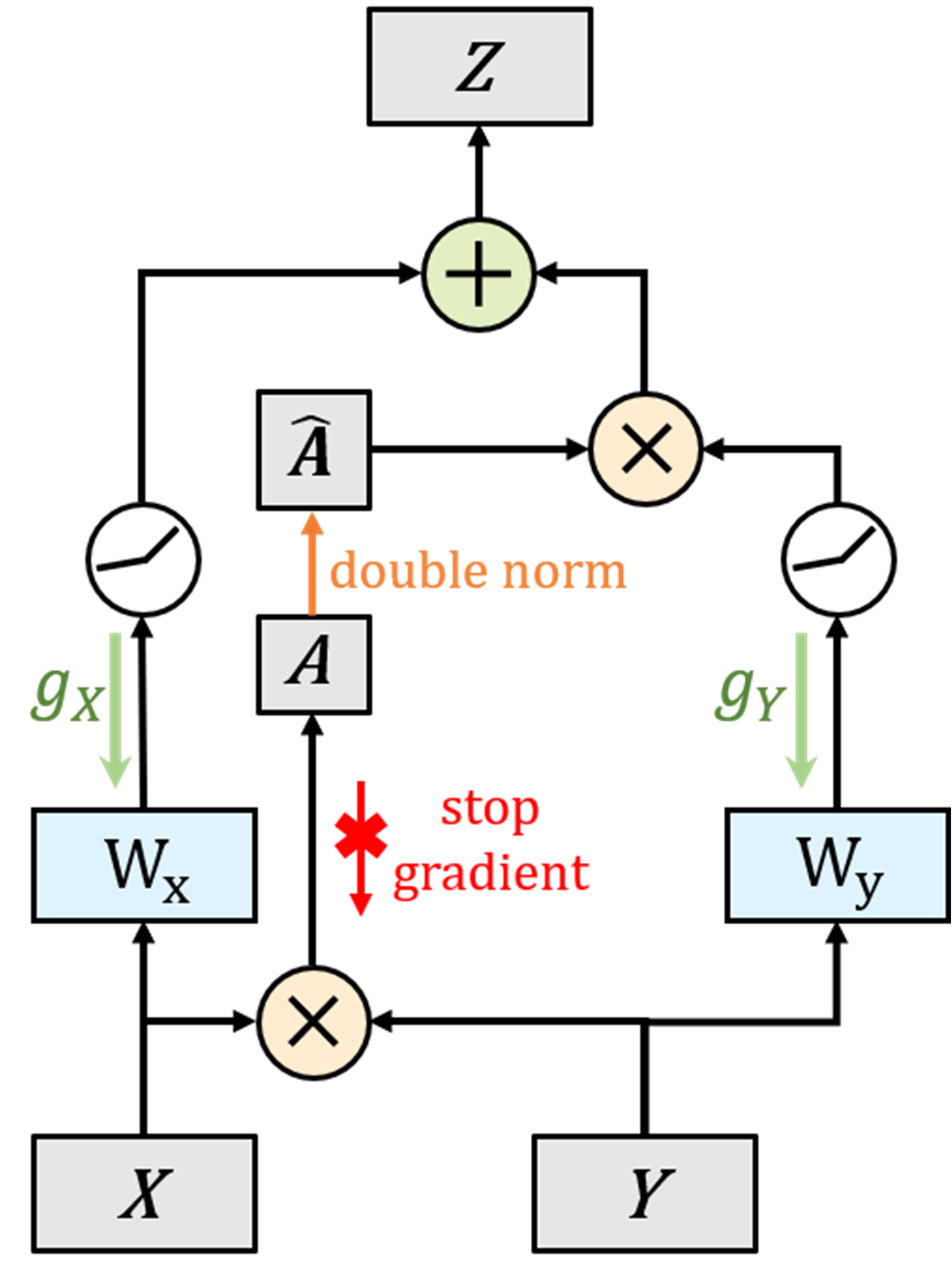}
\caption{SGA}
\label{fig:sga-module}
\end{subfigure}
\begin{subfigure}{0.25\textwidth}
\centering
\includegraphics[width=0.75\linewidth]{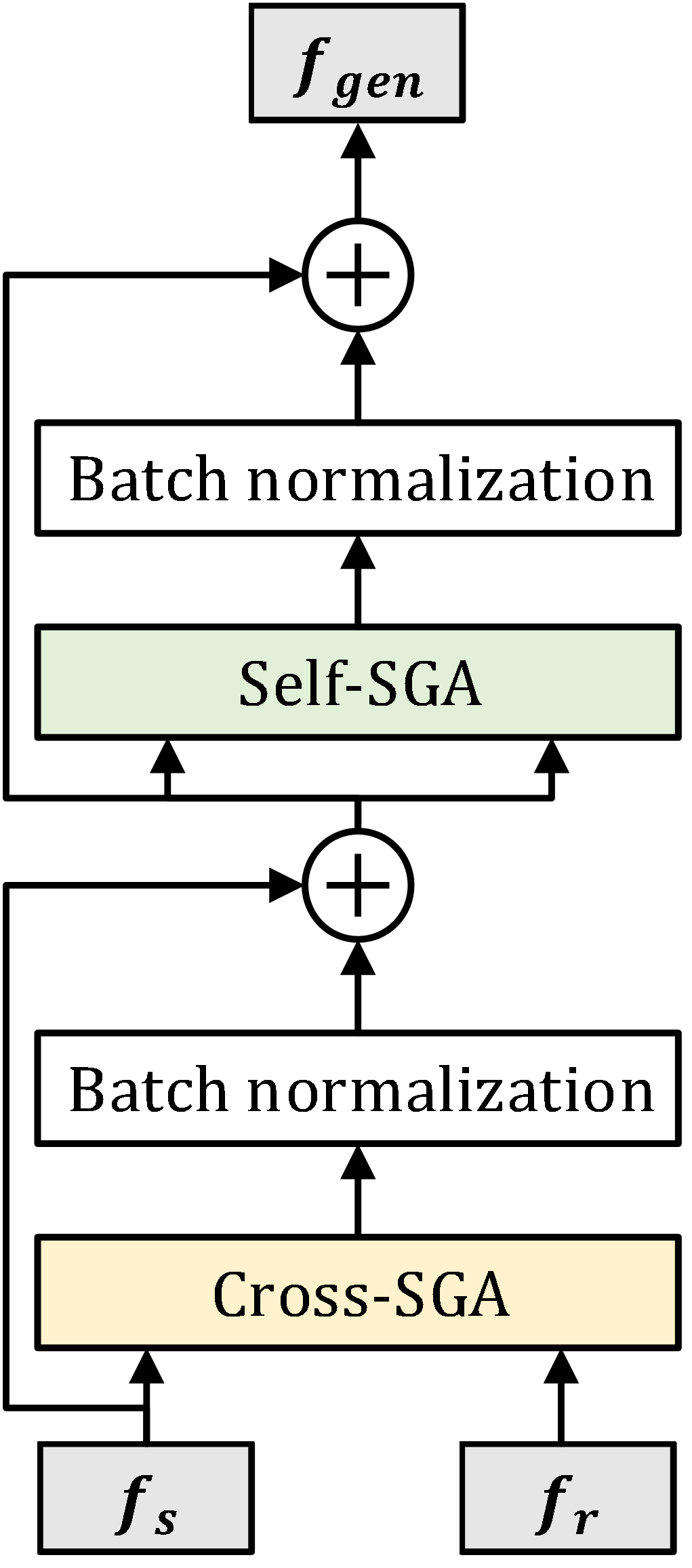}
\caption{SGA Blocks}
\label{fig:sga-blocks}
\end{subfigure}
\caption{
SGA computes the attention map with stop-gradient, which truncates the gradient propagation of $\boldsymbol{g}_{att}$ and adopts a double normalization technique in addition.
In our colorization network, we stack two types of SGA to integrate features: \textcolor[RGB]{191,144,0}{cross-SGA (yellow box)} and \textcolor[RGB]{84,130,53}{self-SGA (green box)}.
}
\label{fig:sgam}
\end{figure}

Combining with the training strategy, we propose the \textbf{S}top-\textbf{G}radient \textbf{A}ttention (SGA).
As \cref{fig:sga-module} illustrates, in addition to the stop-gradient operation, we also design a new feature integration and normalization strategy for SGA.
Treating stop-gradient attention map $\mA$ as a prior deep graph structure input, inspired by \cite{gcn,wang2019learning}, features can be effectively aggregated from adjacency nodes and the node itself:
\begin{equation}
    \mZ = \sigma(\mX \mathbf{W}_{\mathclap{\rm x}}\,) + \widehat{\mA} \sigma(\mY \mathbf{W}_{\mathclap{\rm y}}\,)
    \label{eq:gcn}
\end{equation}
where $\sigma$ is the leaky relu activate function and $\widehat{\mA}$ is the attention map normalized by double normalization method analogous to Sinkhorn algorithm~\cite{cuturi2013sinkhorn}.
Different from softmax employed in classic non-local attention, the double normalization makes the attention map insensitive to the scale of input features~\cite{ex-att}.
The normalized attention map $\hat{\mA}$ can be formulated as follows:
\begin{align}
        \mA &= \mX  \mY^\top  \\
        \Tilde{\mA}_{ij} &= \exp{(\mA_{ij})} / \sum_{k} \exp{(\mA_{ik})}  \\
        \widehat{\mA}_{ij} &= \Tilde{\mA}_{ij} / \sum_{k} \Tilde{\mA}_{kj} 
    \label{eq:graph-construction}
\end{align}
where $\widehat{\mA}_{ij}$ means correlation between $i_{th}$ feature vector in $\mX$ and $j_{th}$ feature vector in $\mY$.
The pseudo-code of SGA is summarized in Algorithm \ref{alg:sga-code}.

\begin{wrapfigure}{L}{0.55\textwidth}
    \begin{minipage}{0.5\textwidth}
      \begin{algorithm}[H]
        \caption{SGA Pseudocode \\ pytorch}
        \label{alg:sga-code}
        \definecolor{codeblue}{rgb}{0.25,0.5,0.5}
        \definecolor{codekw}{rgb}{0.35, 0.35, 0.75}
\lstset{
  language= Python,
  backgroundcolor = \color{white},
  basicstyle      = \fontsize{7.5pt}{7.5pt}\ttfamily\selectfont,
  columns         = fullflexible,
  breaklines      = true,
  captionpos      = b,
  commentstyle    = \fontsize{7.5pt}{7.5pt}\color{codeblue},
  keywordstyle    = \fontsize{7.5pt}{7.5pt}\color{codekw},
  morekeywords    = with,
}
\begin{lstlisting}
# input:
# X: feature maps -> tensor(b, wh, c)
# Y: feature maps -> tensor(b, wh, c)

# output: 
# Z: feature maps -> tensor(b, wh, c)

# other objects:
# Wx, Wy: embedding matrix -> nn.Linear(c,c)
# A: attention map -> tensor(b, wh, wh)
# leaky_relu: leaky relu activation function

with torch.no_grad():
    A = X.bmm(Y.permute(0, 2, 1))
    A = softmax(A, dim=-1)
    A = normalize(A, p=1, dim=-2)
    
X = leaky_relu(Wx(X))
Y = leaky_relu(Wy(Y))

Z = torch.bmm(A,Y) + X
\end{lstlisting}
\end{algorithm}
\end{minipage}
\end{wrapfigure}

Furthermore, we design two types of SGA, called cross-SGA and self-SGA.
Both of their calculation are based on Algorithm~\ref{alg:sga-code}.
As shown in \cref{fig:sga-blocks}, the only difference between them is whether the inputs are the same or not.
Cross-SGA calculates pixel correlation between features from different image domains and integrates features under a stop-gradient attention map.
Self-SGA models the global context and fine-tunes the integration.
For stable training, we also adopt batch normalization layer and shortcut connections~\cite{he2016identity}.
Combining above techniques, our SGA blocks integrate the sketch feature $f_s$ and reference feature $f_r$ into generated feature $f_{gen}$ effectively.

~\\
\section{Experiments}
\subsection{Experiment Setup}
\noindent\textbf{Dataset.}
\quad We test our method on popular anime portraits~\cite{art-editing-dataset} and Animal FacesHQ (AFHQ) \cite{afhq-dataset} dataset. 
The anime portraits dataset contains 33323 anime faces for training and 1000 for evaluation.
AFHQ is a dataset of animal faces consisting of 15,000 high-quality images at 512 × 512 resolution, which contains three categories of pictures, \ie cat, dog, and wildlife.
Each class in AFHQ provides 5000 images for training and 500 for evaluation.
To simulate the line-art drawn by artists, we use XDoG \cite{2012xdog} to extract sketch inputs and set the parameters of XDoG algorithm with $\phi= 1 \times 10^9$ to keep a step transition at the border of sketch lines.
We randomly set $\sigma$ to be 0.3/0.4/0.5 to get different levels of line thickness, which generalizes the network on various line widths to avoid overfitting.
And we set $ p= 19, k = 4.5, \epsilon = 0.01$ by default in XDoG.

~\\
\noindent\textbf{Implementation Details.}
\quad We implement our model with the size of input image fixed at 256×256 for each dataset.
For training, we set the coefficients for each loss terms as follows: 
$\lambda_{1} = 30, \lambda_{2} = 0.01,$ and $\lambda_{3} = 50$.
We use Adam solver \cite{adam} for optimization with $\beta_{1} = 0.5, \beta_{2} = 0.999$.
The learning rate of generator and discriminator are initially set to 0.0001 and 0.0002, respectively.
The training lasts 40 epochs on each dataset. 

\begin{figure*}
\centering
\begin{subfigure}{0.1\textwidth}
\centering
\includegraphics[width=1\linewidth]{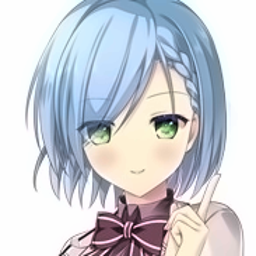}
\includegraphics[width=1\linewidth]{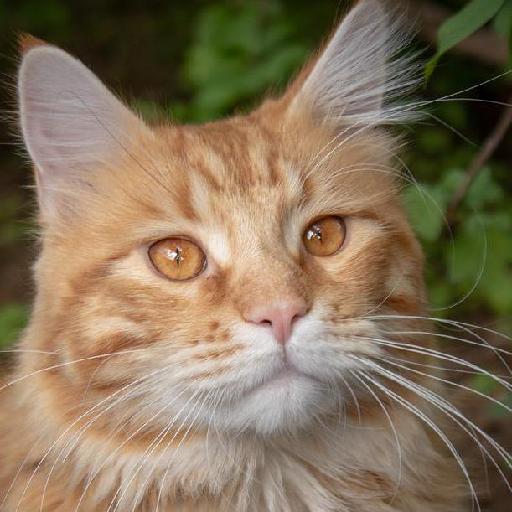}
\includegraphics[width=1\linewidth]{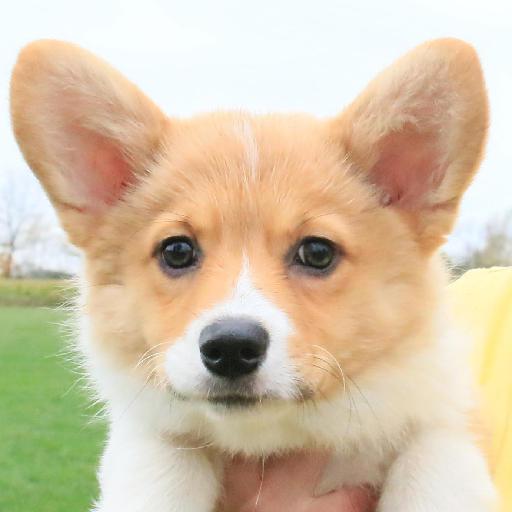}
\includegraphics[width=1\linewidth]{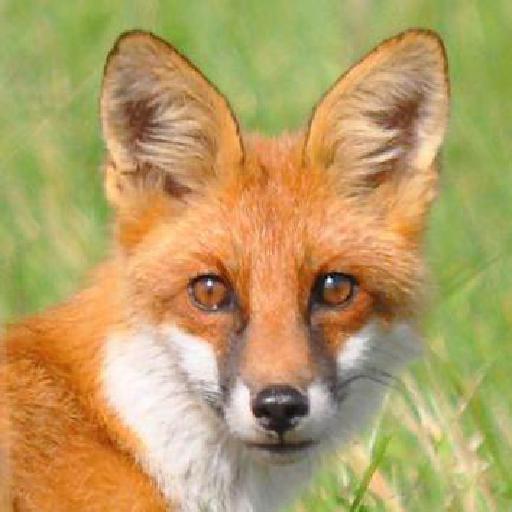}
\caption{Ref}
\end{subfigure}
\begin{subfigure}{0.1\textwidth}
\centering
\includegraphics[width=1\linewidth]{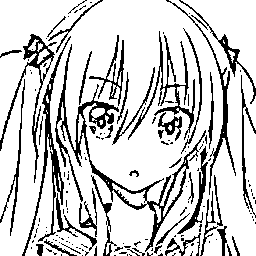}
\includegraphics[width=1\linewidth]{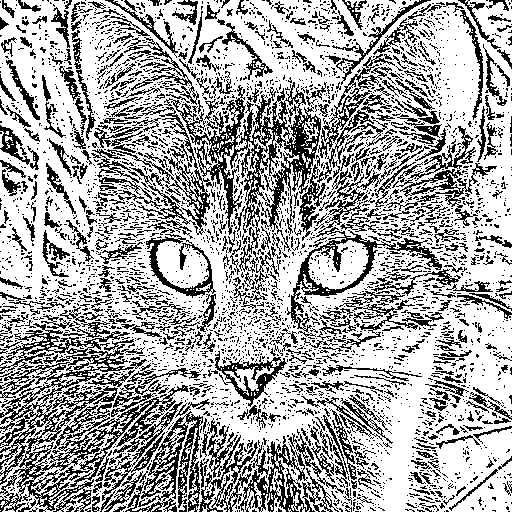}
\includegraphics[width=1\linewidth]{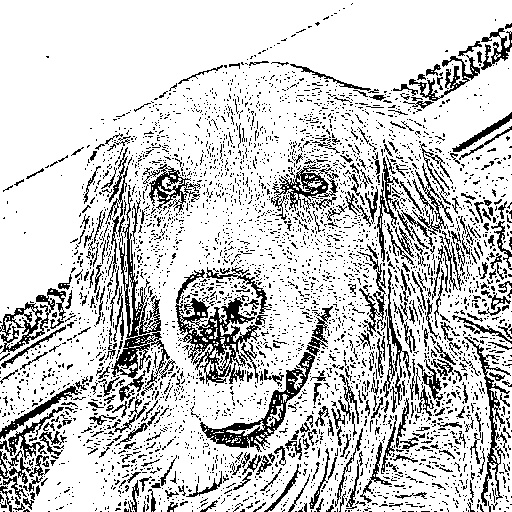}
\includegraphics[width=1\linewidth]{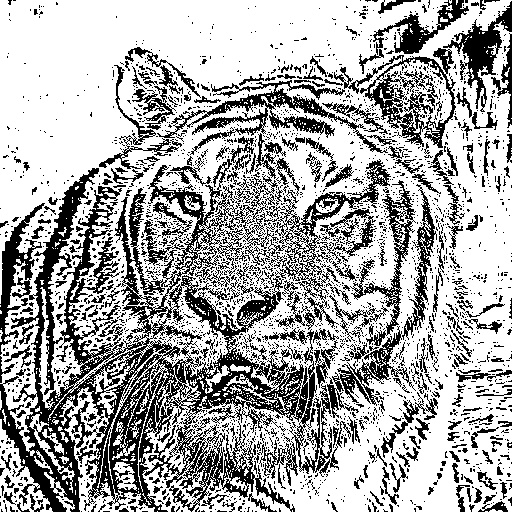}
\caption{Skt}
\end{subfigure}
\begin{subfigure}{0.1\textwidth}
\centering
\includegraphics[width=1\linewidth]{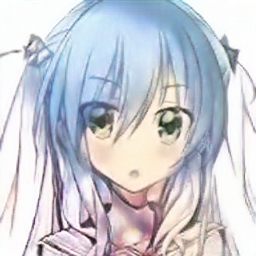}
\includegraphics[width=1\linewidth]{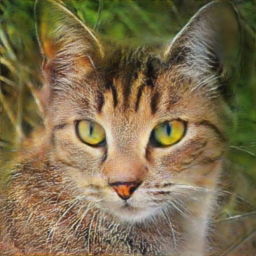}
\includegraphics[width=1\linewidth]{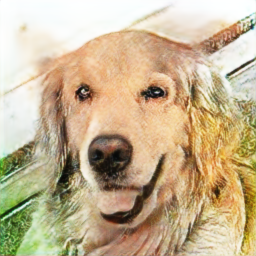}
\includegraphics[width=1\linewidth]{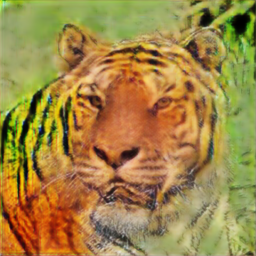}
\caption{\cite{spade}}
\end{subfigure}
\begin{subfigure}{0.1\textwidth}
\centering
\includegraphics[width=1\linewidth]{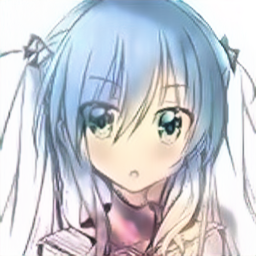}
\includegraphics[width=1\linewidth]{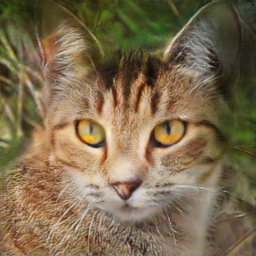}
\includegraphics[width=1\linewidth]{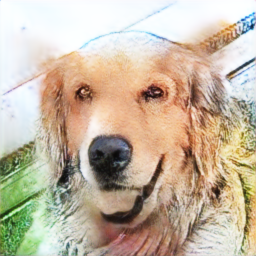}
\includegraphics[width=1\linewidth]{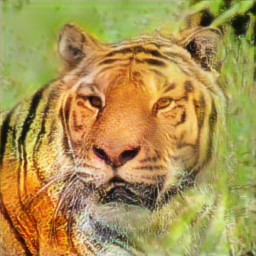}
\caption{\cite{cocosnet}}
\end{subfigure}
\begin{subfigure}{0.1\textwidth}
\centering
\includegraphics[width=1\linewidth]{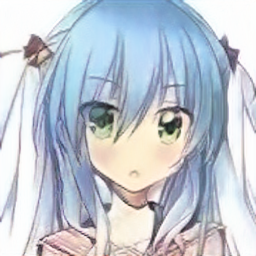}
\includegraphics[width=1\linewidth]{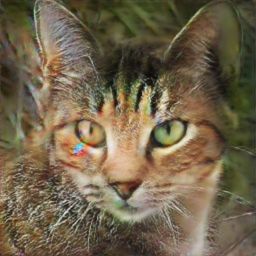}
\includegraphics[width=1\linewidth]{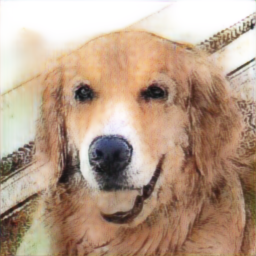}
\includegraphics[width=1\linewidth]{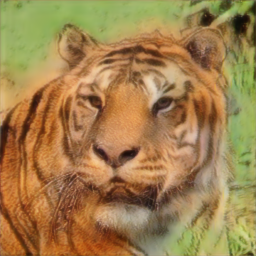}
\caption{\cite{scft}}
\end{subfigure}
\begin{subfigure}{0.1\textwidth}
\centering
\includegraphics[width=1\linewidth]{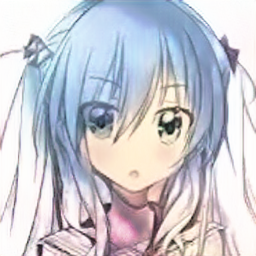}
\includegraphics[width=1\linewidth]{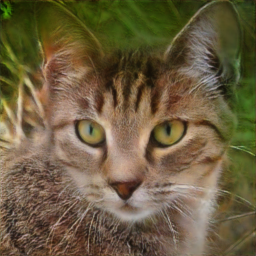}
\includegraphics[width=1\linewidth]{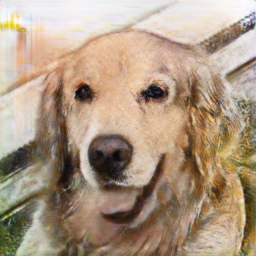}
\includegraphics[width=1\linewidth]{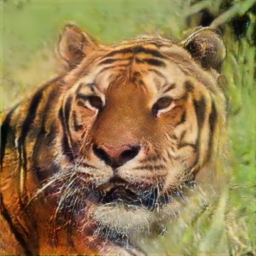}
\caption{\cite{uot-cvpr2021}}
\end{subfigure}
\begin{subfigure}{0.1\textwidth}
\centering
\includegraphics[width=1\linewidth]{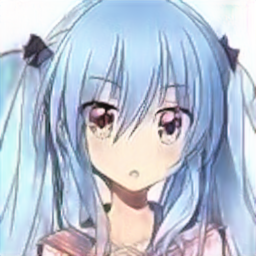}
\includegraphics[width=1\linewidth]{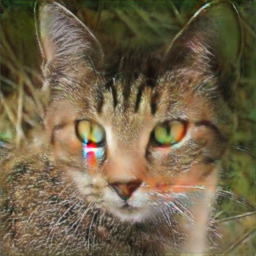}
\includegraphics[width=1\linewidth]{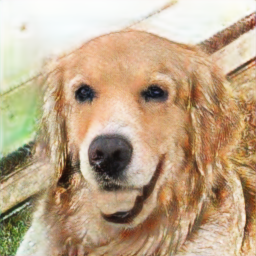}
\includegraphics[width=1\linewidth]{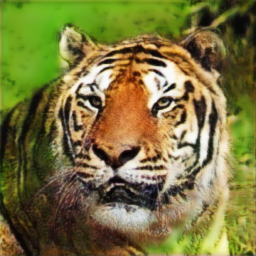}
\caption{\cite{aiqiyi-cmft}}
\end{subfigure}
\begin{subfigure}{0.1\textwidth}
\centering
\includegraphics[width=1\linewidth]{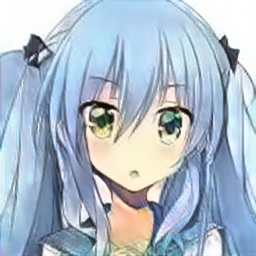}
\includegraphics[width=1\linewidth]{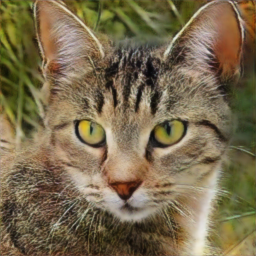}
\includegraphics[width=1\linewidth]{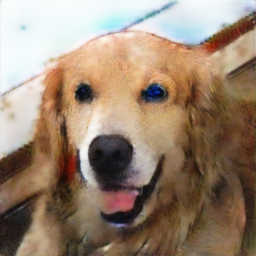}
\includegraphics[width=1\linewidth]{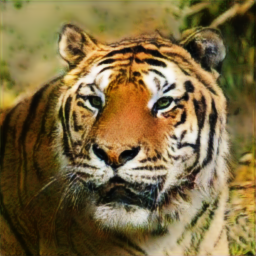}
\caption{Ours}
\end{subfigure}
\caption{
Visualization of colorization results. ``Ref'' stands for ``reference''.``Skt'' indicates ``sketch''.
Compared with other methods, SGA shows correct correspondence between the sketch and reference images.
}
\label{fig:compare-many}
\end{figure*}

~\\
\noindent\textbf{Evaluation Metrics.}
\quad In evaluation process, we randomly select reference images and sketch images for colorization as \cref{fig:compare-many} shows.
The popular Fr$\text{\'{e}}$chet Inception Distance (FID) \cite{fid} is used to assess the perceptual quality of generated images by comparing the distance between distributions of generated and real images in a deep feature embedding.
Besides measuring the perceptual credibility, we also adopt the structural similarity index measure (SSIM) to quantify the outline preservation during colorization, by calculating the SSIM between reference image and original color image of sketch.

\subsection{Comparison Results}\label{sec:cmp}
We compare our method with existing state-of-the-art modules include not only reference-based line-art colorization \cite{scft} but also image-to-image translation, \ie SPADE \cite{spade}, CoCosNet~\cite{cocosnet}, UNITE~\cite{uot-cvpr2021} and CMFT~\cite{aiqiyi-cmft}.
For fairness, in our experiments, all networks use the same encoders, decoder, residual blocks and discriminator implemented in SCFT~\cite{scft} with aforementioned train losses.
Table \ref{tab:result} shows that SGA outperforms other techniques by a large margin.
With respect to our main competitor SCFT, SGA improves by 27.21\% and 25.67\% on average for FID and SSIM, respectively. 
This clear-cut improvement means that SGA produces a more realistic image with high outline preservation compared with previous methods.
According to \cref{fig:compare-many}, the images generated by SGA have less color-bleeding and higher color consistency in perceptual.

\begin{table}[ht]
\centering
\caption{
Quantitative comparison with different methods. 
Boldface represents the best value.
Underline stands for the second score.
}
\begin{tabular}{ccccccccc}
\Xhline{1pt} \specialrule{0em}{0pt}{1pt}
\multirow{2}{*}{Method}       & \multicolumn{2}{c}{anime}      & \multicolumn{2}{c}{cat} & \multicolumn{2}{c}{dog} & \multicolumn{2}{c}{wild} \\ 
\cline{2-9} \specialrule{0em}{0pt}{1.5pt}
 & FID$\downarrow$        & SSIM$\uparrow$         & FID$\downarrow$        & SSIM$\uparrow$        & FID$\downarrow$        & SSIM$\uparrow$        & FID$\downarrow$  & SSIM$\uparrow$ \\ \Xhline{0.75pt} \specialrule{0em}{0pt}{1pt}
SPADE \cite{spade}  &  57.55  & 0.681  &  36.11 &   0.526   & 76.57 &  0.631   &  24.56   &   0.573   \\ \hline \specialrule{0em}{0pt}{1pt}
CoCosNet \cite{cocosnet}  & 52.06 & 0.672 & 35.02 & 0.511 & \underline{68.69} & 0.603 & \underline{23.10} & 0.554  \\ \hline \specialrule{0em}{0pt}{1pt}
SCFT \cite{scft}   &  44.65  & 0.788 & 36.33 & 0.636 & 79.08 & 0.683 & 24.93     &  0.633 \\ \hline \specialrule{0em}{0pt}{1pt}
UNITE \cite{uot-cvpr2021}  & 52.19 & 0.676 & \cellcolor[HTML]{DAE8FC}\textbf{33.26} & 0.636 & 72.38 & 0.677 & 23.97  & 0.592  \\ \hline \specialrule{0em}{0pt}{1pt}
CMFT \cite{aiqiyi-cmft}  & \underline{38.94} & \underline{0.873} & 37.78 & \underline{0.813} & 73.18 & \underline{0.809} & 23.90  & \underline{0.822}  \\ \hline \specialrule{0em}{0pt}{1pt}
SGA & \cellcolor[HTML]{DAE8FC}\textbf{29.65} & \cellcolor[HTML]{DAE8FC}\textbf{0.912} & \underline{34.35} & \cellcolor[HTML]{DAE8FC}\textbf{0.843} & \cellcolor[HTML]{DAE8FC}\textbf{54.76} & \cellcolor[HTML]{DAE8FC}\textbf{0.841} & \cellcolor[HTML]{DAE8FC}\textbf{15.19}  & \cellcolor[HTML]{DAE8FC}\textbf{0.831} \\ 
\Xhline{1pt}
\end{tabular}
\label{tab:result}
\end{table}

\begin{wrapfigure}{r}{0.5\textwidth}
    \centering
    \includegraphics[width=0.9\linewidth]{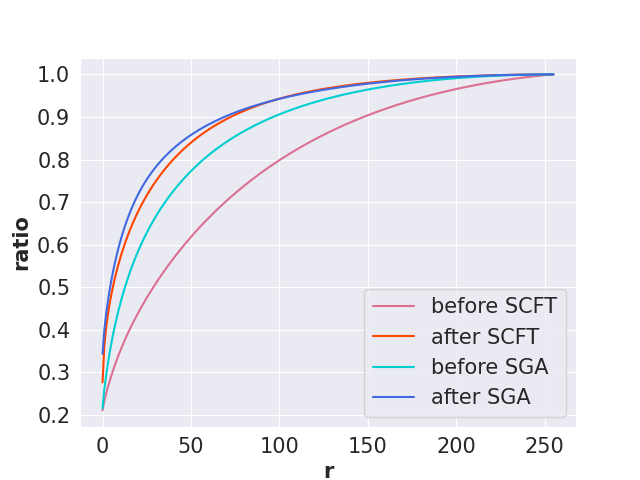}
    \caption{
    Accumulative ratio of the squared top $r$ singular values over total squared singular values in feature maps.
    The ratios of feature maps before and after the attention module in SCFT and SGA are displayed. 
    }
    \label{fig:ratio}
\end{wrapfigure}

Furthermore, we explore the superiority of SGA over SCFT in terms of rescaling spectrum concentration of the representations.
We compare the accumulative ratios of squared top $r$ singular values over total squared singular values of the unfolded feature maps (\ie $\mathbb{R}^{C \times HW}$) before and after passing through the attention module, illustrated in \cref{fig:ratio}.
The sum of singular values is the nuclear norm, \ie the convex relaxation for matrix rank that measures how compact the representations are, which is widely applied in machine learning~\cite{kang2015logdet}.
The accumulative ratios are obviously lifted after going through SCFT and SGA, which facilitates the model to focus more on critical global information~\cite{ham}.
However, our effective SGA can not only further denoise feature maps but also enforce the encoder before attention module to learn energy-concentrated representations, \ie under the effect of SGA, the CNN encoder can also learn to focus on the global information.

\subsection{Ablation Study}\label{sec:ab-study}

We perform several ablation experiments to verify the effectiveness of SGA blocks in our framework, \ie stop-gradient operation, attention map normalization, and self-SGA.
The quantitative results are reported in Table~\ref{tab:sga-ab}, showing the superiority of our SGA blocks.

Specifically, to evaluate the necessity of stop-gradient in non-local attention, we design a variant SGA without stop-gradient.
In Table \ref{tab:sga-ab}, it obtains inferior performance, which verifies the benefit of eliminating gradient conflict through stop-gradient.

\begin{table}[ht]
\centering
\caption{
Ablation study result with different settings.
Boldface represents the best value.
Underline stands for the second best.
}
\begin{tabular}{cclclclcl}
\Xhline{1pt} \specialrule{0em}{0pt}{1pt}
\multirow{2}{*}{Setting} & \multicolumn{2}{c}{anime}  & \multicolumn{2}{c}{cat}    & \multicolumn{2}{c}{dog}    & \multicolumn{2}{c}{wilds}  \\ 
    \cline{2-9} \specialrule{0em}{0pt}{1.5pt}
    & FID$\downarrow$ & SSIM$\uparrow$ & FID$\downarrow$ & SSIM$\uparrow$ & FID$\downarrow$ & SSIM$\uparrow$ & FID$\downarrow$ & SSIM$\uparrow$ \\ 
    \Xhline{0.75pt}  \specialrule{0em}{0pt}{1pt}
SGA & \cellcolor[HTML]{DAE8FC}\textbf{29.65} & \underline{0.912} & \underline{34.35} & \cellcolor[HTML]{DAE8FC}\textbf{0.843} & \cellcolor[HTML]{DAE8FC}\textbf{54.76} & \cellcolor[HTML]{DAE8FC}\textbf{0.841} & \cellcolor[HTML]{DAE8FC}\textbf{15.19}  & \cellcolor[HTML]{DAE8FC}\textbf{0.831} \\ 
\hline   \specialrule{0em}{0pt}{1pt}
SGA w/o stop-gradient & 36.34 & 0.876 & 40.73 & 0.796 & 72.34 & 0.808 & 19.90 & 0.791 \\
\hline   \specialrule{0em}{0pt}{1pt}
SGA w/o double-norm & 33.42 & 0.861 & 34.42 & 0.811 & \underline{55.08} & 0.828 & \underline{15.95} & 0.809 \\
     \hline   \specialrule{0em}{0pt}{1pt}
SGA w/o self-SGA & \underline{31.56} & \cellcolor[HTML]{DAE8FC}\textbf{0.917} & \cellcolor[HTML]{DAE8FC}\textbf{34.26} & \underline{0.842} & 55.69 & \underline{0.839} & 16.36 & \underline{0.821}\\ 
\Xhline{1pt}
\end{tabular}
\label{tab:sga-ab}
\end{table}

Furthermore, we conduct an ablation study on the attention map normalization to validate the advantage of double normalization in our framework.
Table~\ref{tab:sga-ab} demonstrates that SGA with double normalization outperforms that with classic softmax function.
Although classic softmax can generate realistic images, it suffers a low outline preservation, \ie the SSIM measure.

\begin{wrapfigure}{r}{0.4\textwidth}
\centering
\includegraphics[width=0.4\textwidth]{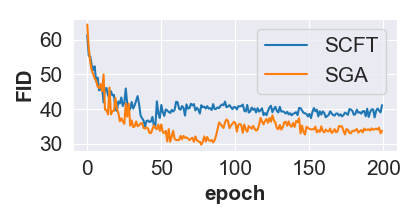}
\caption{Cat's FID during 200 epochs training.}
\label{fig:scft-vs-sga}
\end{wrapfigure}

Based on the framework with stop-gradient and double normalization, we make an ablation study on the improvement of self-SGA additionally. 
Although our model has achieved excellent performance without self-SGA, there is still a clear-cut enhancement on most datasets after employing the self-SGA according to Table~\ref{tab:sga-ab}.
The stacks of SGA can help model not only integrate feature effectively, but also fine-tune a better representation with global awareness for coloring. 

Extending the training schedule to 200 epochs, \cref{fig:scft-vs-sga} shows that SGA can still perform better with more epochs (29.71 in the 78th epoch) and collapse later than SCFT \cite{scft}, demonstrating the training stability for attention models in line-art colorization.

\begin{figure*}
    \centering
    \begin{subfigure}{0.45\textwidth}
    \includegraphics[width=0.9\linewidth]{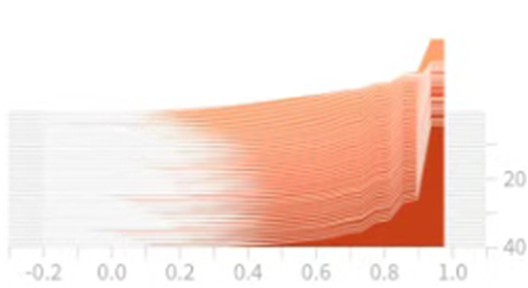}
    \caption{$\cos{(\boldsymbol{g}_{X}, \overline{\boldsymbol{g}_{X}})}$ in branch $\mX$}
    \label{fig:SGA-v-skt}
    \end{subfigure}
    \begin{subfigure}{0.45\textwidth}
    \includegraphics[width=0.9\linewidth]{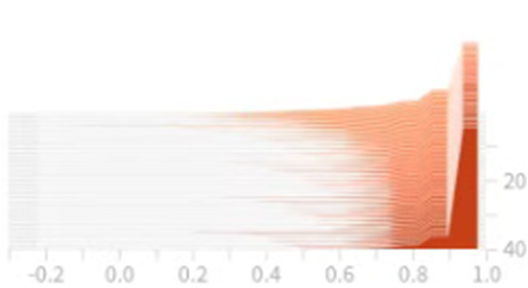}
    \caption{$\cos{(\boldsymbol{g}_{\boldsymbol{Y}},\overline{\boldsymbol{g}_{Y}})}$ in branch $\mY$}
    \label{fig:SGA-v-ref}
    \end{subfigure}
    \begin{subfigure}{0.45\textwidth}
    \includegraphics[width=0.9\linewidth]{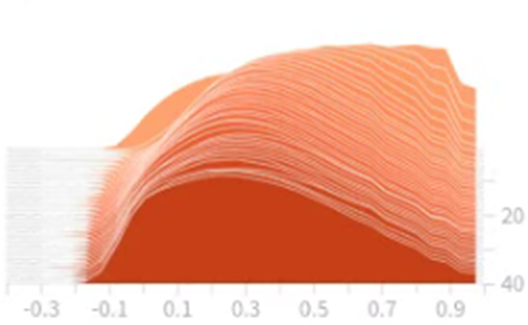}
    \caption{$\cos{(\boldsymbol{g}_{X},\overline{\boldsymbol{g}_{X}}-\boldsymbol{g}_{X})}$ in branch $\mX$}
    \label{fig:SGA-a-skt}
    \end{subfigure}
    \begin{subfigure}{0.45\textwidth}
    \includegraphics[width=0.9\linewidth]{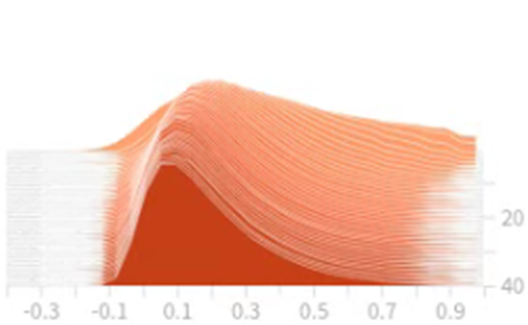}
    \caption{$\cos{(\boldsymbol{g}_{Y},\overline{\boldsymbol{g}_{Y}}-\boldsymbol{g}_{Y})}$ in branch $\mY$}
    \label{fig:SGA-a-ref}
    \end{subfigure}
    \caption{
    The gradient distribution of "SGA w/o stop-gradient". The $\boldsymbol{g}_{X}$ and $\boldsymbol{g}_{Y}$ are illustrated in \cref{fig:sga-module}. The $\overline{\boldsymbol{g}_{X}}$ and $\overline{\boldsymbol{g}_{Y}}$ represent the total gradient, similar to the $\boldsymbol{g}_{skip} + \boldsymbol{g}_{\boldsymbol{Q}}$ and $\boldsymbol{g}_{K} + \boldsymbol{g}_{\boldsymbol{V}}$ in \cref{fig:gradient}.
    }
    \label{fig:SGA-gradient}
\end{figure*}

Additionally, to be more rigorous, we visualize the gradient distributions in the "SGA w/o stop-gradient". \cref{fig:SGA-gradient} implies the existing of gradient conflicts is a general phenomena in dot-product attention mechanism.

\section{Conclusion}

In this paper, we investigate the gradient conflict phenomenon in classic attention networks for line-art colorization. To eliminate the gradient conflict issue, we present a novel cross-modal attention mechanism, \textbf{S}top-\textbf{G}radient \textbf{A}ttention (\textbf{SGA}) by clipping the conflict gradient through the stop-gradient operation.
The stop-gradient operation can unleash the potential of attention mechanism for reference-based line-art colorization. 
Extensive experiments on several image domains demonstrate that our simple technique significantly improves the reference-based colorization performance with better the training stability. 

\noindent\textbf{Acknowledgments:}\,
This research was funded in part by the Sichuan Science and Technology
Program (Nos. 2021YFG0018, 2022YFG0038).

\clearpage
%
%

\clearpage

\appendix

\section{More Results}

\begin{figure}
\centering
\begin{tabular}{c|cccccc}
\diagbox{Skt}{Ref} & \begin{subfigure}{0.12\textwidth}\includegraphics[width=1\linewidth]{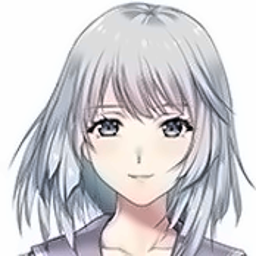}\end{subfigure} & \begin{subfigure}{0.12\textwidth}\includegraphics[width=1\linewidth]{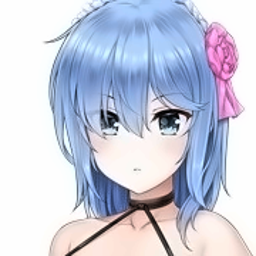}\end{subfigure} & \begin{subfigure}{0.12\textwidth}\includegraphics[width=1\linewidth]{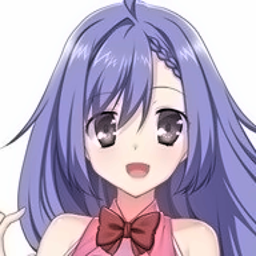}\end{subfigure} & \begin{subfigure}{0.12\textwidth}\includegraphics[width=1\linewidth]{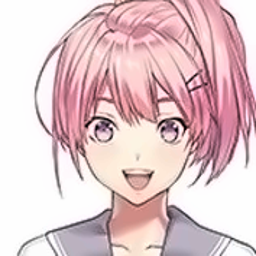}\end{subfigure} & \begin{subfigure}{0.12\textwidth}\includegraphics[width=1\linewidth]{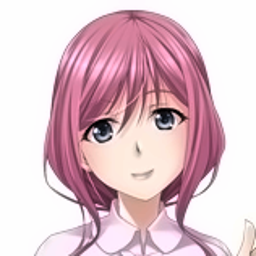}\end{subfigure} & \begin{subfigure}{0.12\textwidth}\includegraphics[width=1\linewidth]{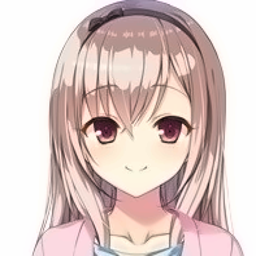}\end{subfigure} \\ \hline
\begin{subfigure}{0.12\textwidth}\includegraphics[width=1\linewidth]{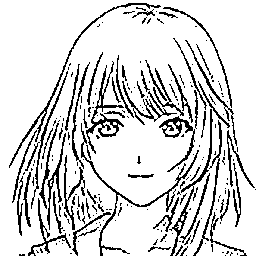}\end{subfigure} & \begin{subfigure}{0.12\textwidth}\includegraphics[width=1\linewidth]{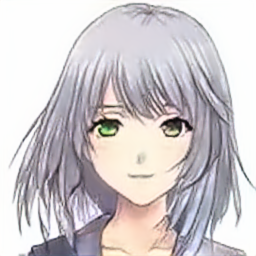}\end{subfigure} & \begin{subfigure}{0.12\textwidth}\includegraphics[width=1\linewidth]{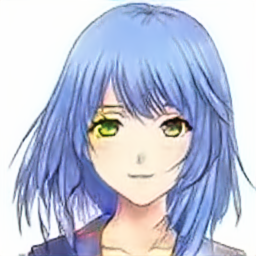}\end{subfigure} & \begin{subfigure}{0.12\textwidth}\includegraphics[width=1\linewidth]{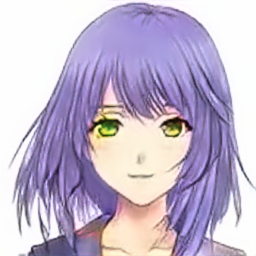}\end{subfigure} & \begin{subfigure}{0.12\textwidth}\includegraphics[width=1\linewidth]{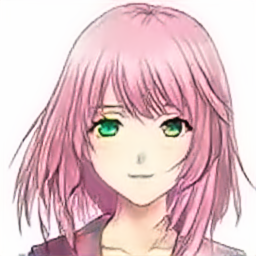}\end{subfigure} & \begin{subfigure}{0.12\textwidth}\includegraphics[width=1\linewidth]{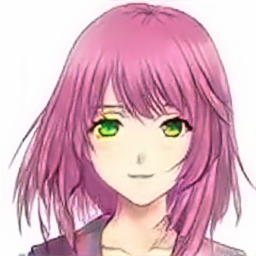}\end{subfigure} & \begin{subfigure}{0.12\textwidth}\includegraphics[width=1\linewidth]{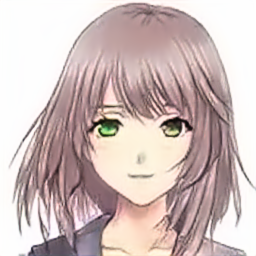}\end{subfigure} \\
\begin{subfigure}{0.12\textwidth}\includegraphics[width=1\linewidth]{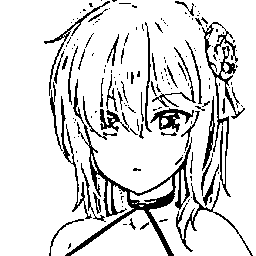}\end{subfigure} & \begin{subfigure}{0.12\textwidth}\includegraphics[width=1\linewidth]{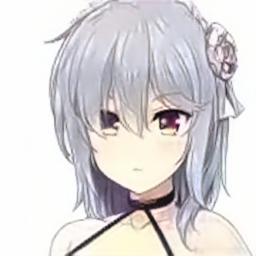}\end{subfigure} & \begin{subfigure}{0.12\textwidth}\includegraphics[width=1\linewidth]{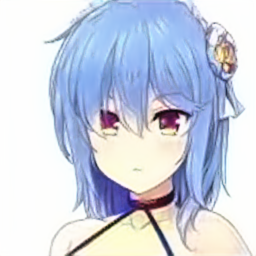}\end{subfigure} & \begin{subfigure}{0.12\textwidth}\includegraphics[width=1\linewidth]{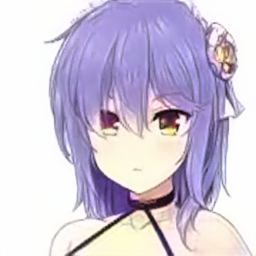}\end{subfigure} & \begin{subfigure}{0.12\textwidth}\includegraphics[width=1\linewidth]{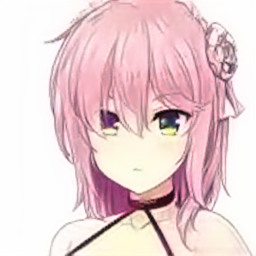}\end{subfigure} & \begin{subfigure}{0.12\textwidth}\includegraphics[width=1\linewidth]{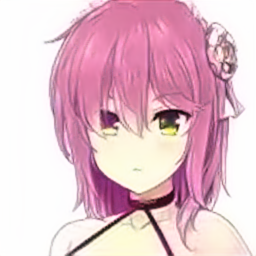}\end{subfigure} & \begin{subfigure}{0.12\textwidth}\includegraphics[width=1\linewidth]{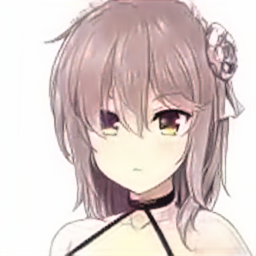}\end{subfigure} \\
\begin{subfigure}{0.12\textwidth}\includegraphics[width=1\linewidth]{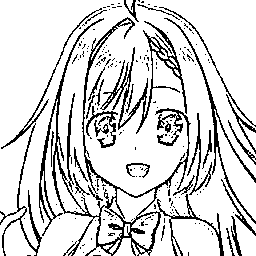}\end{subfigure} & \begin{subfigure}{0.12\textwidth}\includegraphics[width=1\linewidth]{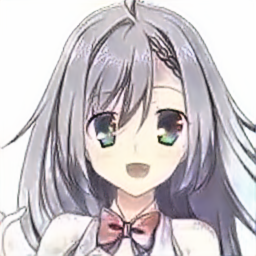}\end{subfigure} & \begin{subfigure}{0.12\textwidth}\includegraphics[width=1\linewidth]{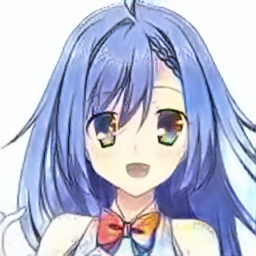}\end{subfigure} & \begin{subfigure}{0.12\textwidth}\includegraphics[width=1\linewidth]{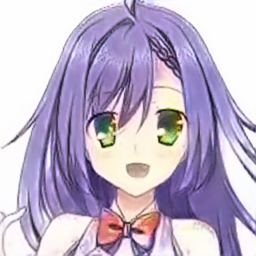}\end{subfigure} & \begin{subfigure}{0.12\textwidth}\includegraphics[width=1\linewidth]{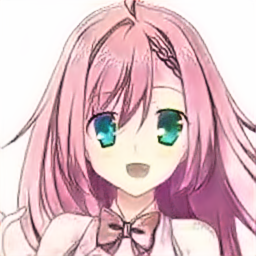}\end{subfigure} & \begin{subfigure}{0.12\textwidth}\includegraphics[width=1\linewidth]{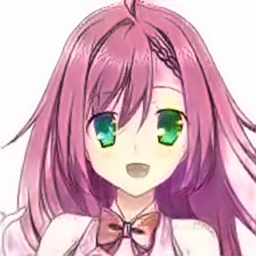}\end{subfigure} & \begin{subfigure}{0.12\textwidth}\includegraphics[width=1\linewidth]{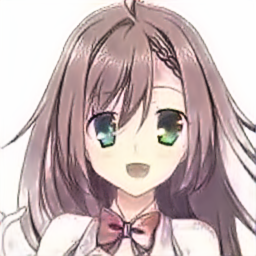}\end{subfigure} \\
\begin{subfigure}{0.12\textwidth}\includegraphics[width=1\linewidth]{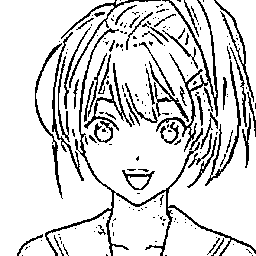}\end{subfigure} & \begin{subfigure}{0.12\textwidth}\includegraphics[width=1\linewidth]{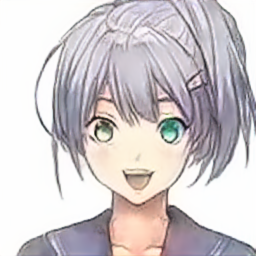}\end{subfigure} & \begin{subfigure}{0.12\textwidth}\includegraphics[width=1\linewidth]{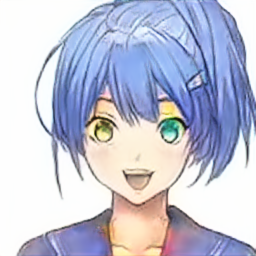}\end{subfigure} & \begin{subfigure}{0.12\textwidth}\includegraphics[width=1\linewidth]{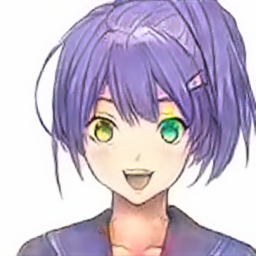}\end{subfigure} & \begin{subfigure}{0.12\textwidth}\includegraphics[width=1\linewidth]{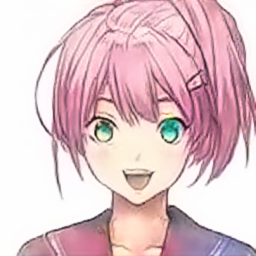}\end{subfigure} & \begin{subfigure}{0.12\textwidth}\includegraphics[width=1\linewidth]{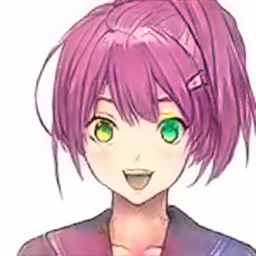}\end{subfigure} & \begin{subfigure}{0.12\textwidth}\includegraphics[width=1\linewidth]{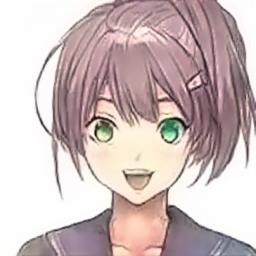}\end{subfigure} \\
\begin{subfigure}{0.12\textwidth}\includegraphics[width=1\linewidth]{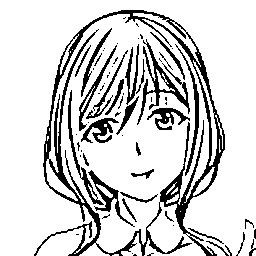}\end{subfigure} & \begin{subfigure}{0.12\textwidth}\includegraphics[width=1\linewidth]{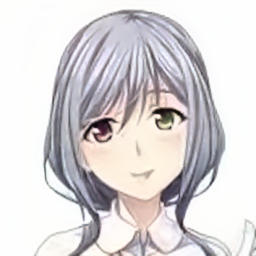}\end{subfigure} & \begin{subfigure}{0.12\textwidth}\includegraphics[width=1\linewidth]{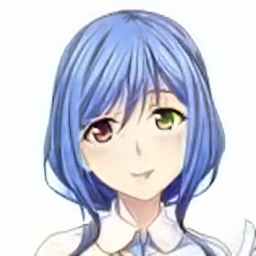}\end{subfigure} & \begin{subfigure}{0.12\textwidth}\includegraphics[width=1\linewidth]{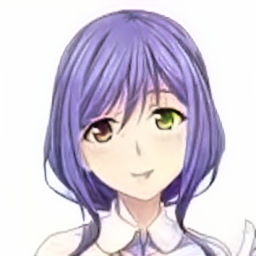}\end{subfigure} & \begin{subfigure}{0.12\textwidth}\includegraphics[width=1\linewidth]{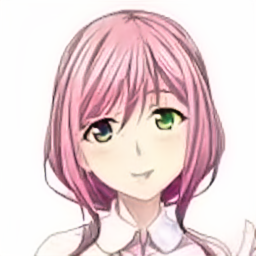}\end{subfigure} & \begin{subfigure}{0.12\textwidth}\includegraphics[width=1\linewidth]{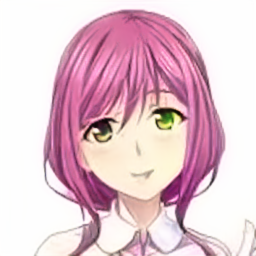}\end{subfigure} & \begin{subfigure}{0.12\textwidth}\includegraphics[width=1\linewidth]{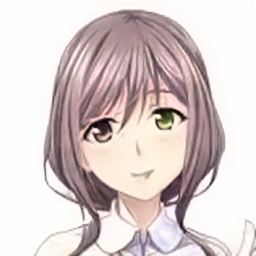}\end{subfigure} \\
\begin{subfigure}{0.12\textwidth}\includegraphics[width=1\linewidth]{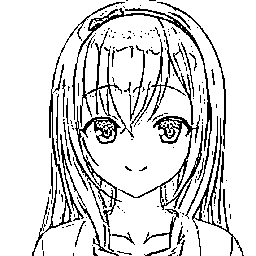}\end{subfigure} & \begin{subfigure}{0.12\textwidth}\includegraphics[width=1\linewidth]{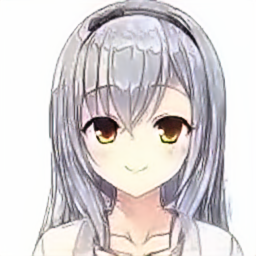}\end{subfigure} & \begin{subfigure}{0.12\textwidth}\includegraphics[width=1\linewidth]{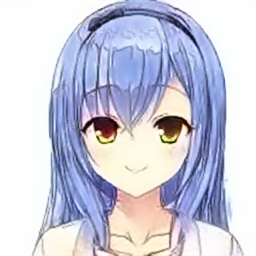}\end{subfigure} & \begin{subfigure}{0.12\textwidth}\includegraphics[width=1\linewidth]{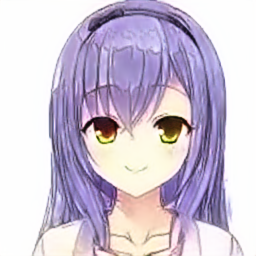}\end{subfigure} & \begin{subfigure}{0.12\textwidth}\includegraphics[width=1\linewidth]{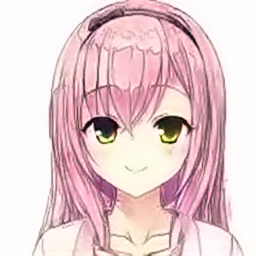}\end{subfigure} & \begin{subfigure}{0.12\textwidth}\includegraphics[width=1\linewidth]{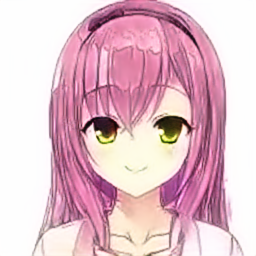}\end{subfigure} & \begin{subfigure}{0.12\textwidth}\includegraphics[width=1\linewidth]{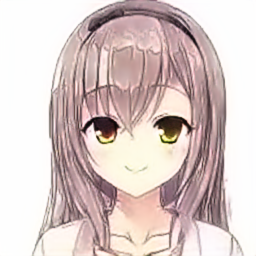}\end{subfigure}
\end{tabular}
\caption{
qualitative results of anime colorization. ``Ref'' stands for ``reference''.``Skt'' indicates ``sketch''.
}
\label{fig:cross}
\end{figure}

To demonstrate the impressive performance of SGA, we add the anime colorization results showed in \cref{fig:cross}.
Not only is the style in reference images appropriately transferred, but also the outline of sketch images are highly persevered, even there existing some divergences of shapes between sketch and reference.

\begin{figure}
\centering
\begin{tabular}{m{1cm}c|c|c|c|c}
&\diagbox{Skt}{Ref} & \begin{subfigure}{0.17\textwidth}\includegraphics[width=1\linewidth]{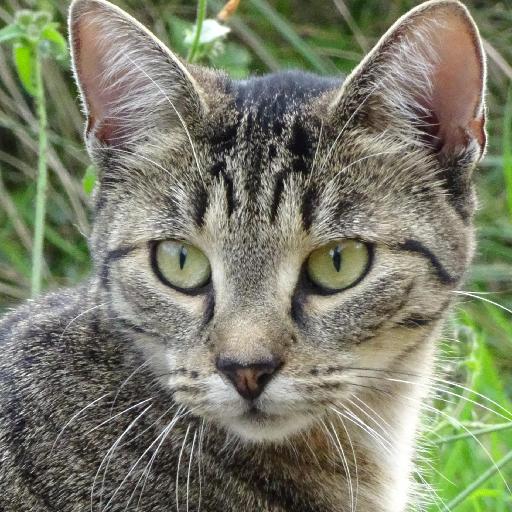}\end{subfigure} & \begin{subfigure}{0.17\textwidth}\includegraphics[width=1\linewidth]{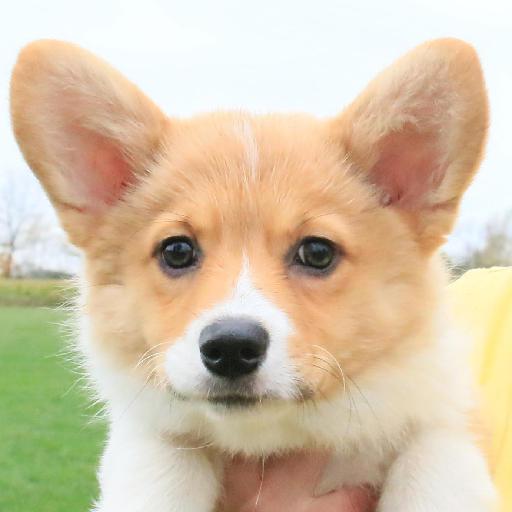}\end{subfigure} & \begin{subfigure}{0.17\textwidth}\includegraphics[width=1\linewidth]{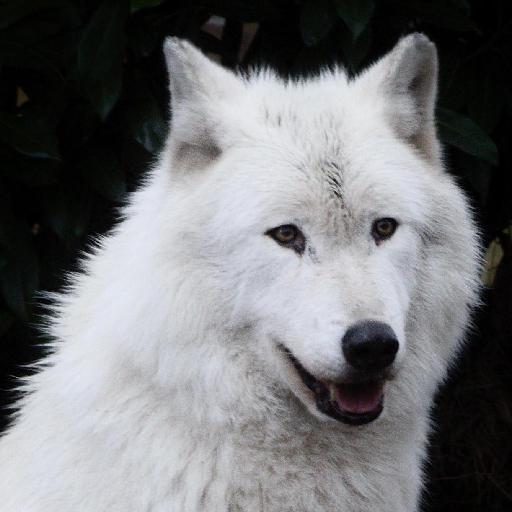}\end{subfigure} & \begin{subfigure}{0.17\textwidth}\includegraphics[width=1\linewidth]{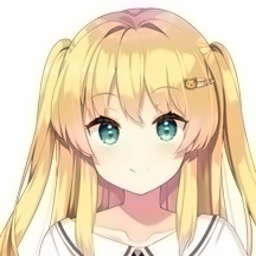}\end{subfigure} \\ \hline
SGA &\begin{subfigure}{0.17\textwidth}\includegraphics[width=1\linewidth]{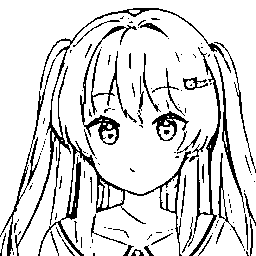}\end{subfigure} & \begin{subfigure}{0.17\textwidth}\includegraphics[width=1\linewidth]{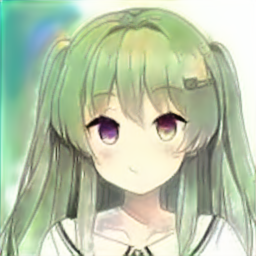}\end{subfigure} & \begin{subfigure}{0.17\textwidth}\includegraphics[width=1\linewidth]{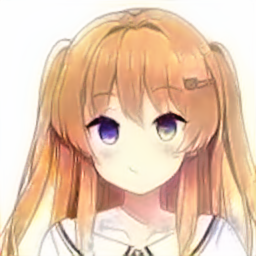}\end{subfigure} & \begin{subfigure}{0.17\textwidth}\includegraphics[width=1\linewidth]{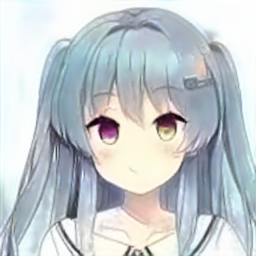}\end{subfigure} & \begin{subfigure}{0.17\textwidth}\includegraphics[width=1\linewidth]{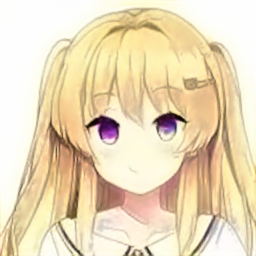}\end{subfigure} \\ \hline
SCFT&\begin{subfigure}{0.17\textwidth}\includegraphics[width=1\linewidth]{supp/ood/skt.png}\end{subfigure} & \begin{subfigure}{0.17\textwidth}\includegraphics[width=1\linewidth]{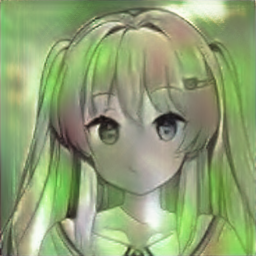}\end{subfigure} & \begin{subfigure}{0.17\textwidth}\includegraphics[width=1\linewidth]{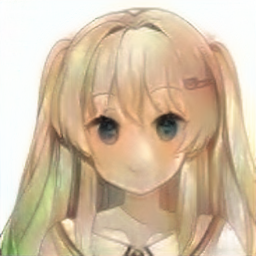}\end{subfigure} & \begin{subfigure}{0.17\textwidth}\includegraphics[width=1\linewidth]{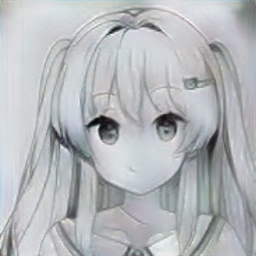}\end{subfigure} & \begin{subfigure}{0.17\textwidth}\includegraphics[width=1\linewidth]{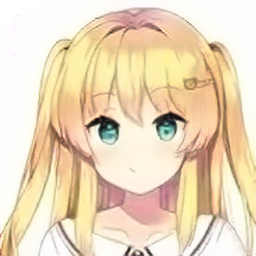}\end{subfigure} \\ \hline
\end{tabular}
\caption{
Visualization of colorization results with out-of-domain reference input, which indicates SGA has a better generalization than SCFT. ``Ref'' stands for ``reference''.``Skt'' indicates ``sketch''.
}
\label{fig:ood}
\end{figure}

Additionally, we find that SGA preforms not bad when facing some huge semantic gap between reference and sketch, through conducting an extreme case, \ie use an out-of-domain reference input to test the generalization in \cref{fig:ood}.
We use the SGA pretrained on anime dataset.
The results show that SGA has a better generalization than SCFT due to the correct style transferring and high outline preservation in this case.

\begin{figure}
    \centering
    \includegraphics[width=0.45\textwidth]{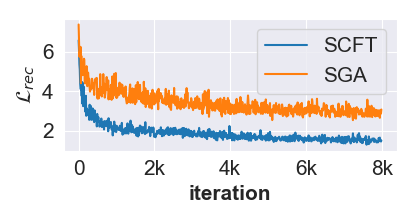}
    \caption{The $\mathcal{L}_{rec}$ during the training process on anime dataset.}
    \label{fig:rec}
\end{figure}

Besides, the last column of \cref{fig:ood} shows that SCFT preform extremely well in self-reconstruction, which implies SCFT is suffered from tending to learn a trivial solution.
The $\mathcal{L}_{rec}$ during the training process showed in \cref{fig:rec} suggests that the SGA has a higher reconstruction loss compared with SCFT.
These evidences shows that the stop-gradient operation helps the model to attain a generalization solution, similar to the SimSiam. 

\end{document}